\documentclass[sn-basic,Numbered]{sn-jnl}

\usepackage{graphicx}%
\usepackage{multirow}%
\usepackage{amsmath,amssymb,amsfonts}%
\usepackage{amsthm}%
\usepackage{mathrsfs}%
\usepackage[title]{appendix}%
\usepackage{xcolor}%
\usepackage{textcomp}%
\usepackage{manyfoot}%
\usepackage{booktabs}%
\usepackage{float}%
\usepackage{algorithm}%
\usepackage{algorithmicx}%
\usepackage{algpseudocode}%
\usepackage{subcaption}%
\usepackage{url}%

\theoremstyle{thmstyleone}%
\theoremstyle{thmstyletwo}%
\theoremstyle{thmstylethree}%
\newtheorem{definition}{Definition}%

\begin{document}

\title[In-Context Collapse]{In-Context Collapse in Vision-Language Models and How to Mitigate it?}

\author[1]{\fnm{Mohammad} \sur{Rostami}}\email{mrostami@seas.upenn.edu}
\affil[1]{\orgname{Amazon Generative AI Innovation Center}\footnote{This work is independent of the position at Amazon}, \orgaddress{\country{USA}}}

\abstract{Many-shot in-context learning (ICL) lets vision-language models (VLMs) adapt from
image--label demonstrations without weight updates, and is widely assumed to improve as more
demonstrations are supplied. We show the opposite is possible: as demonstrations accumulate, a subset of VLMs
undergo an \emph{in-context collapse}, a sharp, sometimes catastrophic drop in accuracy, in some
models to far below chance, spanning synthetic classification, natural-image classification, and
standard visual-question-answering benchmarks. The collapse is a genuine failure of judgment, not of
output formatting: it occurs in-distribution and while the model still emits well-formed labels. Across
an open VLM panel ($0.5$B--$11$B parameters,
several connector designs) and, strikingly, a frontier model (Claude\,Sonnet\,4.5), the collapse is
\emph{graded} rather than universal. Central to this work, two capabilities usually conflated turn out
to be \emph{dissociable}: \emph{robustness} to accumulating demonstrations and the \emph{ability to
learn a novel rule in context}. Their combinations yield three reproducible regimes, which we separate
with contamination-free synthetic concepts that distinguish genuine in-context learning from retrieval
of pretraining priors. We then move from diagnosis to mechanism and remedy. In MLP-projector VLMs, a
parameter-matched lesion-and-rescue causally localizes the collapse to the vision--language
\emph{integration} pathway: a small adapter on the connector and early/mid layers removes it and
restores genuine learning (remap accuracy $0.39\!\rightarrow\!0.91$ at $16$ shots on a
contamination-free task), whereas an \emph{equal-capacity} adapter on the late \emph{readout} layers
does not, and can drive learning below chance; the same integration-not-readout ordering, and the repair
built on it, reproduce across three architectures spanning two connector families. Building on this locus
we propose \textsc{CircA}, an integration-circuit adaptation framework
whose core is a one-time integration \emph{vaccine}: a single adapter, trained once on one synthetic
task, confers collapse-resistance that \emph{transfers} to unseen task families
(chance\,$\rightarrow$\,$0.71$/$0.60$ on held-out CIFAR/Fashion at $16$ shots), while an equal-capacity
late-locus variant fails. A final dissociation carries a continual-learning consequence: the layers
best for in-context \emph{integration} (early/mid) are \emph{not} the layers best for durable
weight-based \emph{consolidation} across a task stream (late), which achieve higher accuracy and less
forgetting at a fraction of the parameters. The in-context collapse is thus an \emph{integration}
failure at the vision--language interface, mechanistically distinct from both the readout and the
consolidation locus, and correctable by a lightweight, transferable intervention. Code and data to
reproduce all results are available at
\url{https://github.com/rostami-m/incontext-collapse}.}


\maketitle

\section{Introduction}\label{sec:intro}
In-context learning (ICL) has become the default way to adapt large models to new tasks. Instead of
collecting a dataset and running gradient descent, a practitioner places a few labeled examples in the
prompt and the model generalizes to a new query, no weight updates, no optimizer, nothing beyond a
forward pass \cite{brown2020gpt3}. For text the capability is now foundational, and as context windows
have grown, few-shot prompting has become \emph{many-shot} prompting, with hundreds or thousands of
demonstrations rivaling fine-tuning \cite{agarwal2024manyshot}. Vision-language models (VLMs) inherit
this promise and raise the stakes: image labels are expensive, visual domains drift after deployment,
and many-shot multimodal ICL is the natural weight-free way to adapt a model to a new domain, label
set, or visual style, attractive precisely because it updates no weights and therefore cannot, by
construction, cause catastrophic forgetting \cite{alayrac2022flamingo,jiang2024manyshotmm}. If accuracy
grew with the number of demonstrations, as it largely does for text, then showing a VLM more examples
would be close to a free lunch.

It is not. We find that supplying more demonstrations frequently makes VLMs \emph{worse}, and often
abruptly. Consider Qwen2-VL-2B classifying CIFAR images into four categories: with no demonstrations it
is correct $94\%$ of the time, yet as correctly labeled examples accumulate its accuracy falls to
$38\%$ by $K{=}64$ shots, and it does so while still emitting well-formed labels, so the drop is a
failure of \emph{judgement}, not of output formatting. Other models fail harder still: LLaVA-OneVision-0.5B
slides from $72\%$ to \emph{below} the $25\%$ of random guessing. The demonstrations did not merely fail
to help; they destroyed a capability the model already had. We call this phenomenon the \emph{in-context
collapse}: as in-context demonstrations accumulate, accuracy falls, sometimes catastrophically and
sometimes below chance. It is easy to miss, because it hides behind reported averages: it emerges only
past a few shots, varies sharply from model to model, and coexists with a second, quieter failure: even
VLMs that \emph{resist} collapse often cannot \emph{learn} a genuinely novel rule from the
demonstrations, silently falling back on pretraining priors instead of the in-prompt mapping. We observe
the collapse across procedurally generated concepts, natural-image classification, and standard
visual-question-answering benchmarks, in open models from $0.5$B to $11$B parameters and, strikingly, in a
frontier model.

Three observations in the recent literature each touch part of this picture but have remained
disconnected. First, several empirical studies of many-shot multimodal ICL report that it can be
unreliable or actively harmful, accuracy plateaus and then \emph{declines}, models imitate the
surface form of the demonstrations rather than reasoning from their content, and VLMs ``struggle to
make use of a larger number of ICL examples''
\cite{huang2025rethinkmmicl,qin2025camamm,zong2025vlicl}. These report the degradation
\emph{behaviorally}, as imitation, copying, or insensitivity to shot count; what has gone unremarked is
that the drop can fall \emph{below random chance while the model still emits well-formed labels}, a
destruction of existing capability rather than a failure to improve, and that it is only one of two
separable failure axes (below). Second, mechanistic studies of VLMs find that visual
information becomes linearly readable only in the \emph{deep} layers of the language model
\cite{neo2024interpvisual,basu2024entitymodal}. Third, VLMs \emph{forget} catastrophically when their
weights are fine-tuned on a sequence of tasks \cite{zhai2023forgetmllm,chen2024coin}. These findings
live in three separate communities (a behavioral anomaly, an interpretability result, and a
continual-learning problem), and the links among them have gone largely unexamined; in particular, the
unreliability of many-shot adaptation has not been tied to a specific, editable mechanism, much less to
a remedy. Where prior work asks \emph{whether} many-shot multimodal ICL works, we ask \emph{why} it
fails, \emph{where} in the network, and \emph{what to do about it}.

This paper both \emph{reports} a phenomenon and \emph{proposes an
algorithm} that addresses it. We proceed in three steps. \emph{(i) Characterize.} We show that \emph{robustness} to
accumulating demonstrations and the \emph{ability to learn a novel rule in context} are dissociable
axes, using contamination-free synthetic concepts whose answers cannot be retrieved from pretraining
and must be learned in the prompt. \emph{(ii) Localize.} A parameter-matched lesion-and-rescue causally
places the collapse at the vision--language \emph{integration} pathway: a small adapter on the
connector and early/mid layers removes it and restores genuine learning, whereas an equal-capacity
adapter on the late \emph{readout} layers does not. \emph{(iii) Repair.} We turn the locus into a
method, \textsc{CircA} (integration-circuit adaptation), whose core is a one-time integration
\emph{vaccine}, a single adapter, trained once on one synthetic task, that confers collapse-resistance
\emph{transferring} to unseen task families. Because ICL is weight-free, the collapse is a capacity
bound on the fast adaptation path; we further show that the integration locus that repairs it is
\emph{not} the locus where weight-based consolidation across a task stream is most durable, yielding a
concrete rule for when to stop prompting and start consolidating.

\paragraph{Our specific contributions include:}
\begin{enumerate}
\item \textbf{The in-context collapse and its graded regimes} (Sec.~\ref{sec:cliff}). Prior work has
noted that many-shot multimodal ICL can degrade or that models copy rather than reason
\cite{huang2025rethinkmmicl,zong2025vlicl}; we sharpen this into a measured phenomenon, accuracy falling
\emph{below chance while outputs remain well-formed}, graded across a panel from $0.5$B to $11$B and at
frontier scale, and we separate it from mere output degeneration. \emph{Takeaway: many-shot adaptation
is not free; whether it helps or harms is set by the alignment recipe, not by parameter count.}
\item \textbf{A robustness--learning dissociation} (Sec.~\ref{sec:dissoc}). With contamination-free
concepts and a label-shuffle control we separate \emph{surviving} demonstrations from \emph{learning}
from them. \emph{Takeaway: a model can be perfectly robust to added demonstrations yet unable to learn
a novel rule, ``does ICL work?'' is two questions, not one.}
\item \textbf{A causal integration locus} (Sec.~\ref{sec:circuit}). A parameter-matched
lesion-and-rescue makes the collapse appear and disappear on demand. \emph{Takeaway: the collapse is
not diffuse but editable, it lives at the connector and early/mid layers, not at the readout.}
\item \textbf{\textsc{CircA}, a transferable repair} (Sec.~\ref{sec:CircA}). A one-time integration
vaccine immunizes a collapse-prone model, and the immunity transfers to unseen tasks
(chance\,$\rightarrow$\,$0.71$/$0.60$ on held-out CIFAR/Fashion), while an equal-capacity late-locus
variant fails. \emph{Takeaway: knowing where the collapse lives is enough to repair it cheaply and
off-line.}
\item \textbf{Integration $\neq$ consolidation} (Sec.~\ref{sec:consol}). The integration locus differs
from where durable weight-based consolidation across a stream is most effective. \emph{Takeaway: the
place to repair fast in-context adaptation is not the place to write slow, durable memories.}
\end{enumerate}

\section{Related Work}\label{sec:related}

Our study sits at the intersection of three literatures that have largely developed in isolation: the
empirical study of multimodal in-context learning, the mechanistic interpretability of vision-language
models, and continual learning for VLMs. We review each in turn, emphasizing both what is established
and what is missing, and we close by stating precisely where this work sits among them.

\paragraph{Multimodal in-context learning.}
The ability to adapt a VLM from a handful of image--text demonstrations, with no weight update, was
established by Frozen \cite{tsimpoukelli2021frozen}, which showed that a frozen language model prompted
with a learned visual prefix can perform few-shot visual question answering, and by Flamingo
\cite{alayrac2022flamingo}, whose gated cross-attention layers interleave visual features into a frozen
language model and enabled strong few-shot transfer from in-context examples. Otter
\cite{li2023otter} then made the capability an explicit training target, instruction-tuning on a corpus
of interleaved in-context image--text examples (MIMIC-IT) so that \emph{following} demonstrations
becomes a trained behavior rather than an emergent side effect. As
context windows grew, attention turned to the \emph{many-shot} regime, where the literature splits.
On closed frontier models, Jiang et al.\ \cite{jiang2024manyshotmm} report that accuracy improves
roughly log-linearly out to hundreds or thousands of demonstrations, suggesting that ``more is
better.'' On open models a contrary picture has emerged: adding demonstrations is frequently unreliable
or harmful, accuracy plateaus and then \emph{declines}, and models appear to imitate the surface form
of the demonstrations rather than reason from them \cite{huang2025rethinkmmicl,qin2025camamm}. Two
lines of work try to \emph{repair} many-shot MICL, tacitly conceding that the naive form often fails.
Multimodal task vectors \cite{huang2024mmtaskvectors} distill a block of demonstrations into a compact
set of activation edits patched into the forward pass at inference, recovering much of the many-shot
benefit while spending almost no context, evidence that the useful content of the demonstrations is
low-dimensional and relocatable. Link-context learning \cite{tai2024linkcontext} instead restructures
the demonstrations, pairing positively and negatively ``linked'' image--label examples with shared
causal structure so a model can bind a genuinely novel word to a novel visual concept from the prompt
alone. Both treat the degradation as given and engineer around it; we instead ask what makes the naive
form fail, and find that the integration locus they implicitly target can be repaired directly. A complementary benchmark line, VL-ICL Bench \cite{zong2025vlicl}, reports that VLMs ``struggle to
make use of a larger number of ICL examples'' and that zero-shot ability does not predict ICL ability,
the closest prior statement to our robustness--learning split; it establishes the symptom across tasks
but does not separate the two axes as distinct regimes, nor localize a cause. Our closest point of
contact is \cite{huang2025rethinkmmicl}, which argues
that VLMs ``mimic rather than reason'' and that accuracy can degrade with more shots. We share that
starting observation but depart from it in three ways. First, we show the degradation is only one of
\emph{two} separable axes formalized as three regimes (Sec.~\ref{sec:dissoc}), a model can be perfectly
robust to added demonstrations and yet wholly unable to learn a novel rule from them, a distinction
neither \cite{huang2025rethinkmmicl} nor \cite{zong2025vlicl} draws. Second, rather than characterizing the failure behaviorally,
we trace it to a specific, editable circuit at the vision--language interface. Third, we connect that
circuit to the downstream question of where adaptation should be written into weights. Where prior work
asks \emph{whether} many-shot MICL works, we ask \emph{why} it fails, \emph{where} in the network, and
\emph{what follows} for adaptation.

\paragraph{In-context learning over images.}
A separate strand performs ICL in the pixel space rather than over image--label pairs. Visual prompting
defines a task by inpainting a grid that already contains a worked input--output example
\cite{bar2022visualprompt}; generalist in-context models such as Painter \cite{wang2023painter} and
SegGPT \cite{wang2023seggpt} cast dense prediction as image-to-image translation conditioned on an
example pair; and other work studies how to \emph{select} good visual exemplars, finding that the
choice of demonstration strongly affects performance \cite{zhang2023goodvisualicl}. These study a
fundamentally different mechanism, completing an image given an example, whereas we study
\emph{classification} ICL, in which the demonstrations encode a discrete image$\rightarrow$label rule.
That difference is what lets us make the task \emph{contamination-free} and thereby separate genuine
in-context rule learning from retrieval of pretraining priors, a separation the pixel-space setting
does not afford.

\paragraph{Mechanistic interpretability of vision-language models.}
A young but fast-growing literature opens up how VLMs process images internally, and it supplies the
machinery our causal analysis builds on. Palit et al.\ \cite{palit2023vlmechint} adapted causal tracing
to BLIP, the first such tool for a VLM. Neo et al.\ \cite{neo2024interpvisual} applied the logit-lens to
visual tokens and argued for a ``process-then-retrieve'' account in which visual information is refined
locally and becomes decision-relevant only in deeper layers; Kaduri et al.\ \cite{kaduri2024whatsimage}
reach a related conclusion in a broad dissection of where and how VLMs use visual content, reporting
that image tokens are queried selectively and predominantly in the middle-to-late layers, and that much
of the visual signal is summarized into a few register-like positions before the answer is formed. Cohen et
al.\ \cite{basu2024entitymodal} show that image-token information reaches the answer only in deep
layers, producing a cross-modal gap in entity-knowledge extraction relative to text, and Golovanevsky
et al.\ \cite{golovanevsky2024notice} introduce a semantic (rather than Gaussian-noise) image-corruption
pipeline to localize the heads and tokens a VLM depends on. Collectively this work establishes
\emph{where visual information becomes readable}, but it stops at correlational localization: it does
not turn a localization into an intervention that adds or removes a behavioral failure. Supplying
exactly that step, a parameter-matched, bidirectional manipulation tied to a specific behavior, is the
methodological core of our paper, and our result refines the readability picture by distinguishing
where the decision is \emph{read out} from where the demonstrations are \emph{integrated}.

\paragraph{In-context learning mechanisms in language models.}
The text-only community has gone further toward mechanism, and we borrow its hypotheses. Induction
heads were identified as a circuit that copies and completes repeated patterns and whose formation
coincides with the emergence of ICL \cite{olsson2022induction}. Hendel et al.\
\cite{hendel2023taskvectors} showed that ICL compresses a task into a single ``task vector'' in the
residual stream, and Todd et al.\ \cite{todd2024functionvectors} found transplantable ``function
vectors'' that causally transport a task between contexts. These results make it natural to expect that
multimodal ICL is likewise computed by an identifiable, editable circuit, but they operate entirely
within the language stream. Our intervention is the multimodal analogue located at the \emph{modality
boundary}: we localize and edit the integration of \emph{visual} demonstrations at the connector and
early/mid layers, a locus with no counterpart in text-only studies and the one place where the visual
and linguistic streams must be bound.

\paragraph{Continual learning for vision-language models.}
Because in-context adaptation is transient, durable adaptation requires writing to weights, which
re-introduces forgetting. Zhai et al.\ \cite{zhai2023forgetmllm} document severe catastrophic
forgetting when multimodal LLMs are fine-tuned sequentially; CoIN \cite{chen2024coin} establishes a
standard continual-instruction-tuning benchmark and shows that even strong VLMs degrade across a task
stream; and SMoLoRA \cite{zheng2024cvitdual} dissects a \emph{dual} forgetting, of both
instruction-following and visual understanding, mitigated by separable low-rank adapters.
The CLiMB benchmark \cite{srinivasan2022climb} specifically studies continual learning across
vision-and-language tasks, finding that standard CL algorithms reduce forgetting but do not enable
cross-task knowledge transfer; transformer architectures that dynamically expand capacity while
distilling prior knowledge address both challenges \cite{cai2023taskattentive,cai2026dynamic}.
Other mitigations are tailored to the multimodal setting: ModalPrompt \cite{zhu2024modalprompt}
maintains a pool of task-specific prompts and routes between them by image--text similarity, avoiding weight
overwrites altogether; CluMo \cite{cai2025clumo} pairs visual and textual prompt keys via clustering for
continual VQA; while ModelTailor \cite{zhu2024modeltailor} identifies and surgically preserves
the small ``forgetting-sensitive'' subset of parameters most responsible for degradation. These build
on the classical continual-learning toolkit: weight-importance regularization that slows updates to
parameters deemed important for past tasks (EWC \cite{kirkpatrick2017ewc}, SI \cite{zenke2017si}) and
function-space distillation from the previous model (LwF \cite{li2017lwf}); episodic and
dark-experience rehearsal that replays stored exemplars or their logits (GEM/A-GEM
\cite{lopezpaz2017gem,chaudhry2019agem}, iCaRL \cite{rebuffi2017icarl}, DER \cite{buzzega2020der});
generative replay that synthesizes past data with a learned generator \cite{shin2017dgr}; and
parameter-efficient or orthogonal-subspace adaptation \cite{hu2022lora,wang2023olora,sun2020lamol},
surveyed in \cite{delange2022clsurvey,vandeven2022threetypes}. This entire literature asks \emph{how} to consolidate
without forgetting; we contribute an orthogonal and, to our knowledge, previously unremarked
observation about \emph{where}: the layers at which weight-based consolidation across a stream is most
effective are \emph{not} the layers at which in-context demonstrations are integrated, which constrains
how parameter-efficient continual methods ought to be targeted. The broader fast/slow division between
weight-free and weight-based adaptation echoes complementary learning systems in the brain
\cite{mcclelland1995cls,kumaran2016cls,oreilly2014cls} and the in-context vs.\ in-weight interplay
studied in humans and networks \cite{russin2025iclinweight}; cognitively inspired models of incremental
concept learning draw on the same CLS theory to enable networks to expand knowledge without
cross-task interference \cite{rostami2023cognitively}; we use this only as a light organizing
lens, not as a claim.

Three facts are, by now, each established in isolation. (i) Many-shot multimodal ICL can
\emph{degrade}, and degrade in a way that looks like imitation rather than reasoning
\cite{huang2025rethinkmmicl,qin2025camamm}. (ii) Visual information becomes linearly readable only in
the \emph{deep} layers of a VLM \cite{neo2024interpvisual,basu2024entitymodal}. (iii) VLMs
\emph{forget} catastrophically under sequential fine-tuning \cite{zhai2023forgetmllm,chen2024coin}.
Each has been pursued in a separate community (a behavioral anomaly, an interpretability finding, and
a continual-learning problem), and the connections among them have gone largely unremarked. The thesis
of this paper is that they are three views of one underlying structure. The behavioral degradation~(i)
is only one half of a robustness--learning dissociation; we causally localize it to the integration
interface with a parameter-matched lesion-and-rescue, and in doing so we \emph{refine} the readability
picture~(ii) by showing that the late layers \emph{read out} the decision but do not \emph{integrate}
the demonstrations, readability and manipulability come apart. The same intervention reveals that the
integration locus is distinct from where weight-based consolidation~(iii) is most durable, so the
remedy for the collapse and the remedy for forgetting live in different parts of the network. Our
contribution is therefore not a new model, benchmark, or training method, but a unifying causal account
that converts a scattered set of observations into a single, editable mechanism at the
vision--language boundary.

\section{The In-Context Collapse and How to Address It}\label{sec:proposed}

We first establish that many-shot
multimodal ICL can \emph{collapse}, then argue that the collapse is a localized failure of a specific
computation, and finally use that diagnosis to design a fix. This section develops the phenomenon and
the algorithm; the empirical evidence for every claim made here is given in
Sec.~\ref{sec:empirical}. We first fix notation for a VLM and for many-shot multimodal ICL
(Sec.~\ref{sec:prelim}), use it to define the collapse and the robustness--learning dissociation
precisely (Sec.~\ref{sec:proposed-pheno}), formalize the integration/readout partition and the
lesion-and-rescue intervention that localizes the failure (Sec.~\ref{sec:proposed-where}), and finally
state the three components of \textsc{CircA} as operations on that partition
(Sec.~\ref{sec:proposed-CircA}).

\subsection{Preliminaries and notation}\label{sec:prelim}

\paragraph{A vision--language model as a composition of maps.}
We consider an autoregressive VLM that consumes an \emph{interleaved} sequence of images and text and
emits text. It factorizes into four stages. A \emph{vision encoder}
$g_\phi:\mathcal{X}\to\mathbb{R}^{n_v\times d_v}$ maps an image $x\in\mathcal{X}$ to $n_v$ visual feature
vectors. A \emph{connector} $c_\psi:\mathbb{R}^{n_v\times d_v}\to\mathbb{R}^{m\times d}$ projects those
features into the $d$-dimensional token-embedding space of the language model, producing $m$ visual
tokens (for an MLP-projector model $m=n_v$ and $c_\psi$ is a small feed-forward network; for a
resampler/perceiver it is a fixed set of learned queries). A \emph{language model} is a stack of $L$
pre-norm transformer blocks $h^{(\ell)}_{\theta_\ell}:\mathbb{R}^{T\times d}\to\mathbb{R}^{T\times d}$,
$\ell=1,\dots,L$, each acting on a length-$T$ sequence of $d$-dimensional hidden states by the residual
updates
\begin{equation}
\tilde H^{(\ell)} = H^{(\ell-1)} + \mathrm{Attn}^{(\ell)}\!\big(H^{(\ell-1)}\big),
\qquad
H^{(\ell)} = \tilde H^{(\ell)} + \mathrm{MLP}^{(\ell)}\!\big(\tilde H^{(\ell)}\big),
\label{eq:block}
\end{equation}
where $\mathrm{Attn}^{(\ell)}$ and $\mathrm{MLP}^{(\ell)}$ are the attention and feed-forward sublayers,
each a set of linear projections with frozen weights. Finally an \emph{unembedding}
$W_U\in\mathbb{R}^{|\mathcal{V}|\times d}$ maps the last hidden state of the final layer to logits over
the vocabulary $\mathcal{V}$. Writing the full network as $f_\theta$ with parameters
$\theta=(\phi,\psi,\theta_{1:L},W_U)$, the next-token distribution given a token/visual-token sequence
$s$ is $p_\theta(\cdot\mid s)=\mathrm{softmax}\big(W_U\,H^{(L)}_{\text{last}}(s)\big)$.

\paragraph{Tasks, verbalizers, and the \textsc{standard}/\textsc{remap} conditions.}
A \emph{concept task} is a distribution $\mathcal{D}$ over pairs $(x,y)$ with $x\in\mathcal{X}$ an image
and $y\in\mathcal{C}=\{1,\dots,C\}$ a class. A \emph{verbalizer}
$v:\mathcal{C}\to\mathcal{V}^{*}$ assigns each class a label string. The verbalizer is the lever that
separates retrieval from learning, and it defines our two conditions:
\begin{itemize}
\item \textsc{standard}: $v=v_{\mathrm{nat}}$ maps each class to its natural name (e.g.\ $v(\text{cat
class})=\texttt{"cat"}$), so a correct answer may come either from the pretraining prior or from the
demonstrations.
\item \textsc{remap}: $v=\rho$ is a fixed bijection onto \emph{arbitrary} tokens disjoint from the
natural names (e.g.\ class $\mapsto$ \texttt{"blorp"}). Because $\rho$ is independent of pretraining,
above-chance accuracy under \textsc{remap} can \emph{only} arise from binding the in-prompt
mapping, never from retrieval. This is what we mean by \emph{contamination-free}.
\end{itemize}

\paragraph{Many-shot multimodal ICL.}
A $K$-shot episode draws a support set of demonstrations
$S_K=\big((x_1,y_1),\dots,(x_K,y_K)\big)$ and a query $x_q$ i.i.d.\ from $\mathcal{D}$, class-balanced,
and forms the interleaved prompt
\begin{equation}
P_K \;=\; \big(\,x_1,\,v(y_1),\;x_2,\,v(y_2),\;\dots,\;x_K,\,v(y_K),\;x_q\,\big),
\label{eq:prompt}
\end{equation}
each image first mapped to visual tokens by $g_\phi$ and $c_\psi$. The model predicts the class whose
label string is most probable,
$\hat y_q=\arg\max_{c\in\mathcal{C}} p_\theta\!\big(v(c)\mid P_K\big)$, and we report the balanced
accuracy
\begin{equation}
A^{v}(K)\;=\;\Pr_{S_K,x_q\sim\mathcal{D}}\big[\hat y_q=y_q\big],
\label{eq:acc}
\end{equation}
as a function of the shot count $K$, separately for $v=v_{\mathrm{nat}}$ (written $A^{\mathrm{std}}$) and
$v=\rho$ (written $A^{\mathrm{rmp}}$). Chance accuracy is $1/C$. The $K{=}0$ point is the zero-shot
prior: $A^{\mathrm{std}}(0)$ measures pretrained competence on the task, while
$A^{\mathrm{rmp}}(0)\approx 1/C$ (chance) by construction, since the arbitrary labels carry no
pretraining signal and a model with no demonstrations can do no better than guess among them.

\subsection{The phenomenon}\label{sec:proposed-pheno}
\begin{figure}[!ht]
\centering
\includegraphics[width=0.62\textwidth]{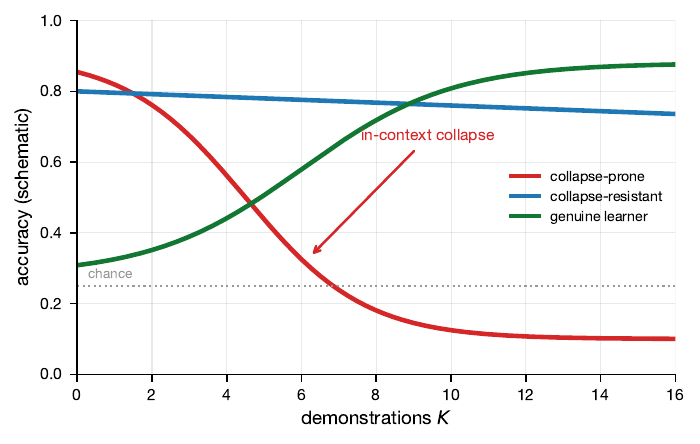}
\caption{\textbf{The in-context collapse (schematic).} As demonstrations accumulate, a
\emph{collapse-prone} model's accuracy falls, often below chance, while a \emph{collapse-resistant}
model is flat and a \emph{genuine learner} improves. Robustness (does accuracy survive added
demonstrations?) and learning (does accuracy rise on a novel rule?) are separable axes whose
combinations define three regimes, established empirically in Sec.~\ref{sec:empirical}.}
\label{fig:phenomenon}
\end{figure}

By the \emph{in-context collapse} we mean a simple but counter-intuitive behavior: for a subset of VLMs,
accuracy on a task \emph{decreases} as in-context demonstrations are added, sometimes gradually, often
abruptly, and in the worst cases to below chance (Fig.~\ref{fig:phenomenon}, red). With the notation of
Sec.~\ref{sec:prelim} in hand, we can state this more accurately. Whereas the canonical few-shot intuition is
that accuracy is non-decreasing in $K$, the collapse is the opposite monotonicity.

\begin{definition}[In-context collapse]\label{def:collapse}
A model exhibits an \emph{in-context collapse} on a task under verbalizer $v$ if its accuracy curve
$A^{v}(K)$ from Eq.~\eqref{eq:acc} is decreasing past some shot count: there exist $K_1<K_2$ with
$A^{v}(K_2) < A^{v}(K_1)$. We call the collapse \emph{catastrophic} when
$A^{v}(K_2)\le A^{v}(0)$ (the demonstrations leave the model worse than seeing none) and
\emph{sub-chance} when $A^{v}(K_2) < 1/C$. We summarize its magnitude by the drop
$\delta^{v} = \max_{K} A^{v}(K) - A^{v}(K_{\max})$.
\end{definition}

The collapse is not universal: some VLMs hold a flat curve as demonstrations accumulate (blue) or even
improve (green). That gradedness is precisely what makes it dangerous in practice: a practitioner who
validates many-shot prompting on one model and deploys it on another can silently lose most of the
model's accuracy. Crucially, two capabilities that are usually conflated come apart, and the contrast
between the two verbalizers makes them separately measurable.

\begin{definition}[Robustness and learning]\label{def:axes}
For a task and model we define two scalar axes from the accuracy curves of Eq.~\eqref{eq:acc}:
\emph{robustness}
$\mathrm{Rob} = A^{\mathrm{std}}(K_{\max}) - A^{\mathrm{std}}(0)$, the change in
\textsc{standard} accuracy as demonstrations accumulate (negative under collapse); and \emph{learning}
$\mathrm{Lrn} = A^{\mathrm{rmp}}(K_{\max}) - A^{\mathrm{rmp}}(0)$, the gain
in \textsc{remap} accuracy over its at-chance starting point, which by contamination-freeness measures the
ability to acquire a novel in-prompt rule (any above-chance \textsc{remap} accuracy can come only from the
demonstrations).
\end{definition}

Robustness and learning are logically independent: a model may be perfectly robust ($\mathrm{Rob}\approx
0$) yet unable to learn ($\mathrm{Lrn}\approx 0$), answering from its prior rather than the
demonstrations, or it may learn while collapsing on the natural labels. Thresholding the two axes
partitions models into the regimes of Fig.~\ref{fig:phenomenon} (made empirical in
Sec.~\ref{sec:cliff}--\ref{sec:dissoc}): \emph{collapse-prone} ($\mathrm{Rob}\!\ll\!0$),
\emph{collapse-resistant retrieval} ($\mathrm{Rob}\!\approx\!0,\ \mathrm{Lrn}\!\approx\!0$), and
\emph{genuine learner} ($\mathrm{Rob}\!\gtrsim\!0,\ \mathrm{Lrn}\!\gg\!0$). The collapse is thus graded
and recipe-dependent rather than a function of scale, and the \textsc{remap} verbalizer $\rho$ is what
lets above-chance accuracy be attributed to learning the in-prompt mapping, never to retrieval.

\subsection{Why it happens, and where to intervene}\label{sec:proposed-where}
\begin{figure}[!ht]
\centering
\includegraphics[width=\textwidth]{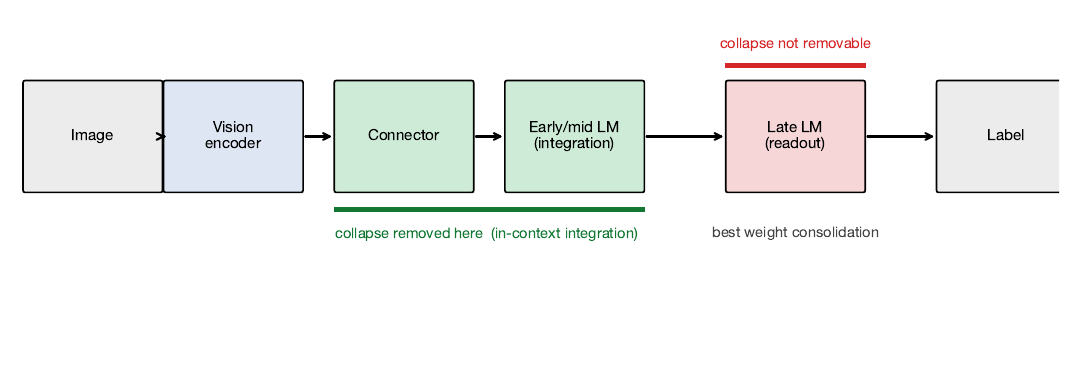}
\caption{\textbf{The in-context collapse is an integration failure at the vision--language
interface.} A VLM maps an image through a vision encoder and a connector into a language model whose
early/mid layers \emph{integrate} the in-context demonstrations and whose late layers \emph{read out}
the decision. We show the collapse is causally removable by a small adapter on the connector and
early/mid layers (green) but not by an equal-capacity adapter on the late readout (red); conversely,
durable weight-based consolidation across a task stream is most effective at that same late
readout, a double dissociation between the integration and consolidation loci.}
\label{fig:concept}
\end{figure}

A VLM routes an image through a vision encoder and a connector into a language model whose early/mid
layers must \emph{integrate} the interleaved demonstrations and whose late layers \emph{read out} the
answer (Fig.~\ref{fig:concept}). We make ``where'' precise by partitioning the model's parameters into
disjoint loci by depth.

\paragraph{The integration/readout partition.}
Index the connector together with the $L$ language-model blocks, and split the block indices into
contiguous thirds, $\mathcal{E}=\{1,\dots,\lfloor L/3\rfloor\}$ (early),
$\mathcal{M}=\{\lfloor L/3\rfloor{+}1,\dots,\lfloor 2L/3\rfloor\}$ (mid), and
$\mathcal{L}=\{\lfloor 2L/3\rfloor{+}1,\dots,L\}$ (late). We group these into an \emph{integration}
locus and a \emph{readout} locus,
\begin{equation}
\mathcal{I} \;=\; \{c_\psi\}\cup\mathcal{E}\cup\mathcal{M},
\qquad
\mathcal{R} \;=\; \mathcal{L},
\label{eq:partition}
\end{equation}
where $\mathcal{I}$ is the pathway that binds visual demonstrations to their in-prompt labels at the
modality boundary, and $\mathcal{R}$ is the pathway that maps an already-formed decision to output
tokens. Our central hypothesis is mechanistic and falsifiable.

\begin{definition}[Integration hypothesis]\label{def:hyp}
The in-context collapse is a deficit of the \emph{integration} computation localized to $\mathcal{I}$,
not of the readout $\mathcal{R}$: as demonstrations accumulate, the vision--language integration
pathway is overwhelmed and the model reverts to its prior instead of binding the in-prompt mapping.
\end{definition}

\paragraph{Lesion-and-rescue as a causal test.}
To test Definition~\ref{def:hyp} we ask whether adding a small amount of trainable capacity \emph{at a
chosen locus} can restore learning that the frozen model lacks. Concretely, for a target region
$\mathcal{G}\subseteq\mathcal{I}\cup\mathcal{R}$ we attach a low-rank adapter \cite{hu2022lora} to the
linear projections in $\mathcal{G}$: each such weight $W\in\mathbb{R}^{d_{\text{out}}\times d_{\text{in}}}$
is replaced by $W+\Delta_{\mathcal{G}}$ with $\Delta_{\mathcal{G}}=\tfrac{\alpha}{r}BA$,
$B\in\mathbb{R}^{d_{\text{out}}\times r}$, $A\in\mathbb{R}^{r\times d_{\text{in}}}$, rank
$r\ll\min(d_{\text{in}},d_{\text{out}})$, while \emph{all} original parameters $\theta$ stay frozen.
We then train only $\Delta_{\mathcal{G}}$ on \textsc{remap} episodes, minimizing the query
cross-entropy
\begin{equation}
\Delta^\star_{\mathcal{G}} \;=\; \arg\min_{\Delta_{\mathcal{G}}}\;
\mathbb{E}_{S_K,x_q\sim\mathcal{D}}\Big[-\log p_{\theta\oplus\Delta_{\mathcal{G}}}\big(\rho(y_q)\mid P_K\big)\Big],
\label{eq:rescue}
\end{equation}
and re-measure the rescued learning curve $A^{\mathrm{rmp}}_{\mathcal{G}}(K)$. A region \emph{rescues}
the collapse if it lifts learning well above the unintervened baseline,
$A^{\mathrm{rmp}}_{\mathcal{G}}(K_{\max}) \gg A^{\mathrm{rmp}}(K_{\max})$. The hypothesis predicts an
asymmetry: rescue should succeed when $\mathcal{G}\subseteq\mathcal{I}$ and fail when
$\mathcal{G}=\mathcal{R}$.

\paragraph{Controlling for capacity.}
Because a larger adapter could rescue \emph{trivially} by sheer added capacity, the test is only
meaningful at matched budget. We therefore compare an integration region against the readout at an
\emph{equal module count} $|\mathcal{G}_{\mathcal{I}}|\approx|\mathcal{G}_{\mathcal{R}}|$ and equal rank
$r$ (e.g.\ on Qwen2-VL-2B the early and late regions differ by under $10\%$ of adapter modules; see
Sec.~\ref{sec:setup}). Under this control, any difference between rescuing at $\mathcal{I}$ versus
$\mathcal{R}$ reflects \emph{location}, not parameter count, so a confirmed asymmetry,
\begin{equation}
A^{\mathrm{rmp}}_{\mathcal{I}}(K_{\max}) \;\gg\; A^{\mathrm{rmp}}(K_{\max})
\;\approx\; A^{\mathrm{rmp}}_{\mathcal{R}}(K_{\max}),
\label{eq:asym}
\end{equation}
is a causal localization of the collapse to the integration pathway. Sec.~\ref{sec:circuit} confirms
Eq.~\eqref{eq:asym} with a parameter-matched lesion-and-rescue, turning the hypothesis about ``where''
into a causal handle we build on.

\subsection{\textsc{CircA}: repairing the integration circuit}\label{sec:proposed-CircA}
The diagnosis in Sec.~\ref{sec:proposed-where} is constructive: it does not merely say that the collapse
exists, it says \emph{where} the fault lies (the integration locus $\mathcal{I}$) and \emph{what} would
fix it (more capacity placed exactly there). \textsc{CircA} (integration-circuit adaptation) is the
operationalization of that diagnosis.

\begin{figure}[!ht]
\centering
\includegraphics[width=\textwidth]{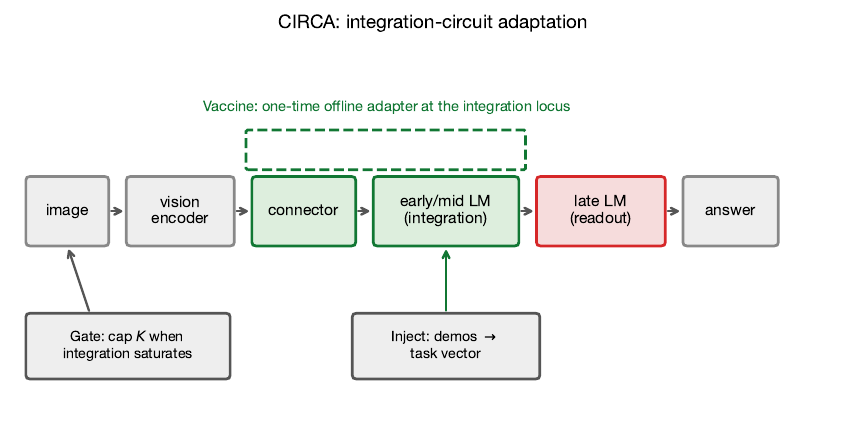}
\caption{\textbf{\textsc{CircA}: integration-circuit adaptation.} A VLM maps an image through a vision
encoder and connector into a language model that first \emph{integrates} demonstrations (early/mid,
green) and then \emph{reads out} the answer (late, red). \textsc{CircA} targets the integration locus
with three complementary components: a one-time offline \emph{vaccine} (adapter on the
connector\,+\,early/mid layers), an inference-time \emph{gate} that caps the demonstration budget at
saturation, and an \emph{inject} path that adds a demonstration-derived task vector at the same locus.}
\label{fig:CircA_arch}
\end{figure}

A collapse-prone model has the right perceptual machinery but a \emph{weak integration step}: when many
demonstrations arrive, it fails to bind them to the query and falls back on copying the most recent
label or answering from its pretraining prior. \textsc{CircA} addresses this in two complementary ways.
First, if we are allowed a small one-time offline investment, we \emph{strengthen} the integration step
directly: we train a tiny adapter at $\mathcal{I}$ on a single synthetic task and obtain a model that
now integrates demonstrations in general, a procedure we call a \emph{vaccine} because it is administered
once, ahead of time, and confers lasting resistance. Second, if we must take the model as given at
serving time, we \emph{avoid} the regime where integration breaks: a \emph{gate} stops adding
demonstrations once a cheap diagnostic shows the model has started copying, and an \emph{inject} path
folds extra demonstrations into a single activation edit instead of spending context on them. The three
components share one locus, $\mathcal{I}$, and split cleanly by what they assume: the vaccine
\emph{creates} integration capacity where it is missing, whereas the gate and inject \emph{exploit}
integration capacity that already exists (Fig.~\ref{fig:CircA_arch}). We now define each as an explicit
operation on the model.

\subsubsection{The vaccine: a transferable integration adapter.}
The vaccine is the load-bearing component and is the same intervention as the diagnostic
lesion-and-rescue of Sec.~\ref{sec:proposed-where} (Eq.~\eqref{eq:rescue}), used now as a remedy rather
than a probe. Recall that
in the diagnosis we showed a small low-rank adapter placed at the integration locus $\mathcal{I}$ can
\emph{restore} a model's ability to learn an arbitrary in-context rule. The vaccine asks the natural
follow-up question: is the restored ability \emph{specific} to the task we trained on, or have we
installed a \emph{general} ``read the demonstrations and use them'' skill that carries over to new tasks?
If the latter, the fix becomes practical: train once, deploy everywhere.

Concretely, we attach a low-rank adapter $\Delta_{\mathcal{I}}$ to the integration locus, freeze every
original weight $\theta$, and train \emph{only} $\Delta_{\mathcal{I}}$ on a \emph{single}
contamination-free task $\mathcal{T}_0$ in the \textsc{remap} condition (arbitrary labels $\rho$, so the
only way to be right is to use the demonstrations). The training objective is ordinary supervised
fine-tuning, minimizing the negative log-likelihood of the query's correct (remapped) label given the
$K$-shot prompt,
\begin{equation}
\Delta^\star \;=\; \arg\min_{\Delta_{\mathcal{I}}}\;
\mathbb{E}_{S_K,x_q\sim\mathcal{T}_0}\Big[-\log p_{\theta\oplus\Delta_{\mathcal{I}}}\big(\rho(y_q)\mid P_K\big)\Big],
\qquad
f^{\,\mathrm{vac}} := f_{\theta\oplus\Delta^\star}.
\label{eq:vaccine}
\end{equation}
In other words, across many sampled demonstration sets $S_K$ and queries $x_q$ from the one training task, push
the model to predict the query's true label from the in-prompt mapping alone. The result is a single
immunized model $f^{\,\mathrm{vac}}$, which is then \emph{frozen and reused} on every downstream task with
no further training, no per-task tuning, and no labels at deployment. Three design choices are
deliberate and worth making explicit, because they are what make the vaccine a one-time, transferable
fix rather than ordinary task-specific fine-tuning. \emph{(i)~Locus.} The adapter is confined to
$\mathcal{I}$ (connector and early/mid layers), the site the diagnosis identified; placing the same
capacity at the readout does not work (the control of Eq.~\eqref{eq:asym}). \emph{(ii)~A
contamination-free training task.} Training on \textsc{remap} arbitrary labels prevents the adapter from
memorizing a label$\rightarrow$answer shortcut and forces it to learn the demonstration-using
\emph{procedure}, which is what transfers. \emph{(iii)~Parameter efficiency.} $\Delta_{\mathcal{I}}$ is a
rank-$r$ update on a small subset of modules ($\sim\!1\%$ of parameters), so the vaccine is cheap to
train and adds negligible inference cost. The claim, tested in Sec.~\ref{sec:CircA}, is that
$\Delta^\star$ installs a general capability rather than a $\mathcal{T}_0$-specific lookup, so that for
unseen task families $\mathcal{T}\neq\mathcal{T}_0$ the learning curve satisfies
$A^{\mathrm{rmp}}_{f^{\,\mathrm{vac}}}(K_{\max};\mathcal{T}) \gg A^{\mathrm{rmp}}_{f_\theta}(K_{\max};\mathcal{T})$,
i.e.\ a model that could not learn novel rules in context now can, on tasks it was never trained
on.\footnote{We use \emph{vaccine} in the sense of a one-time offline intervention conferring
transferable resistance. The term has been used independently for harmful-fine-tuning-robust alignment
\cite{huang2024vaccine} and, as \emph{inoculation}, for prompt-based suppression of undesired behaviors
\cite{wichers2025inoculation}; our usage and mechanism (an integration-locus adapter against in-context
collapse) are unrelated to both.}

\subsubsection{The gate: stop before saturation.}
The gate addresses a different situation: we cannot modify the model (no offline access, a closed API),
so we cannot vaccinate it. The only lever left is \emph{how many demonstrations we feed it}. Since
accuracy falls as $K$ grows past a model-specific point, the safe policy is to stop adding
demonstrations just before that point, getting the benefit of a few shots without paying the cost of
collapse. The difficulty is detecting the onset of collapse without ground-truth labels at serving time.
We use a cheap, label-free signal, the \emph{copy rate}, defined as the fraction of queries whose
prediction equals the label attached to the \emph{most recent} demonstration:
\begin{equation}
\hat r(K) \;=\; \Pr_{S_K,x_q}\big[\hat y_q = y_K\big].
\label{eq:copyrate}
\end{equation}
Intuitively, a healthy model's answer should depend on the query image, not on which demonstration
happened to come last; if the prediction tracks the last label instead, the model has stopped reasoning
and started copying. A rising $\hat r(K)$ is therefore the behavioral fingerprint of the collapse, and,
crucially, it needs no labels to compute, only the model's own outputs. Given a tolerance $\tau$, the
gate serves with the largest shot count whose copy rate is still acceptable,
\begin{equation}
K^\star \;=\; \max\{\,K : \hat r(K)\le\tau\,\},
\label{eq:gate}
\end{equation}
truncating the demonstration budget at the onset of saturation. The copy rate is estimated once on a
small held-out probe, so the gate is essentially free. Unlike the vaccine, it adds no parameters and
changes nothing inside the model; it simply avoids operating the model in the regime where its
integration step is known to fail.

\subsubsection{The inject path: amortize context at the integration locus.}
The third component targets cost rather than accuracy. Demonstrations are expensive: each image consumes
hundreds of tokens, so a many-shot prompt can be slow and may not even fit in the context window. The
inject path lets a model that can \emph{already} integrate pay that cost once and reuse the result. The
idea, the multimodal analogue of a task/function vector in language models
\cite{hendel2023taskvectors,todd2024functionvectors,huang2024mmtaskvectors}, is that the useful effect of
a block of demonstrations on the model's internal state is low-dimensional and can be summarized by a
single vector. We compute that vector by measuring how a demonstration block $D$ shifts the model's
hidden state at the integration locus, relative to seeing no demonstrations at all:
\begin{equation}
v_{\mathcal{I}}(D) \;=\; \tfrac{1}{|\mathcal{I}_\star|}\!\!\sum_{\ell\in\mathcal{I}_\star}\!\!
\big(\bar H^{(\ell)}(D) - \bar H^{(\ell)}(\varnothing)\big),
\label{eq:inject}
\end{equation}
Reading the equation: for each chosen integration layer $\ell\in\mathcal{I}_\star$, $\bar H^{(\ell)}(D)$
is the hidden state (at a fixed reference position) when the prompt contains the demonstrations and
$\bar H^{(\ell)}(\varnothing)$ is the same quantity with no demonstrations; their difference is the
``demonstration effect'' at that layer, and we average it over layers to obtain one vector
$v_{\mathcal{I}}(D)$. At inference we then answer the query with $v_{\mathcal{I}}(D)$ \emph{added back}
into the residual stream at those layers, reproducing the effect of the demonstrations while spending
\emph{no} demonstration tokens in the prompt. The key conceptual point, which also delimits the
framework, is that injection can only \emph{relocate} a computation the model already performs: if the
model cannot integrate demonstrations in the first place, the extracted vector encodes nothing useful.
This is exactly why inject amortizes context for a vaccinated or collapse-resistant model but cannot, by
itself, repair a collapse-prone one, a prediction we confirm in Sec.~\ref{sec:CircA}.

Our experiments load-test this division of labor (Sec.~\ref{sec:CircA}): the vaccine of
Eq.~\eqref{eq:vaccine} confers collapse-resistance that \emph{transfers} to unseen task families; a
locus ablation shows the repair must live at $\mathcal{I}$, since an equal-capacity adapter at the
readout $\mathcal{R}$ fails entirely (the asymmetry of Eq.~\eqref{eq:asym}); and the training-free
injection of Eq.~\eqref{eq:inject}, which can only relocate a computation the model already performs,
does not by itself repair a collapse-prone model, marking the boundary of the framework.

\paragraph{Which component to use when?}
The three components are not alternatives to choose among but a small decision tree keyed to deployment
constraints. If one can touch the model offline even once, apply the \emph{vaccine}: it is the only
component that raises the accuracy ceiling on a collapse-prone model, and it is paid for once. If the
model is fixed at serving time (e.g.\ a closed API), use the \emph{gate} to stay out of the collapsed
regime. If, in addition, prompt length or latency is a binding constraint and the model can already
integrate (natively or post-vaccine), use \emph{inject} to amortize the demonstrations into a single
edit. Vaccine and gate/inject compose: a vaccinated model can still be gated and injected at serving
time.

Algorithm~\ref{alg:CircA} states the framework in full and mirrors this split. Lines~1--4 are the
one-time offline vaccine that produces the immunized model $f^{\,\mathrm{vac}}$; lines~5--11 are the
per-deployment serving policy, in which the gate caps the shot budget at saturation
(line~6) and the inject path optionally amortizes a block of demonstrations into a single task-vector
edit (lines~7--9) when context budget is tight.

\begin{algorithm}[t]
\caption{\textsc{CircA}: integration-circuit adaptation}
\label{alg:CircA}
\begin{algorithmic}[1]
\Require VLM $f_\theta$; integration/readout partition $\mathcal{I},\mathcal{R}$ (Eq.~\eqref{eq:partition});
contamination-free synthetic task $\mathcal{T}_0$ with \textsc{remap} verbalizer $\rho$; copy-rate
threshold $\tau$.
\Statex \textit{// One-time offline vaccine (Eq.~\eqref{eq:vaccine})}
\State Build \textsc{remap} prompts $P_K$ from $\mathcal{T}_0$ (arbitrary labels $\rho$; query supervised by $\rho(y_q)$)
\State Attach low-rank adapter $\Delta_{\mathcal{I}}$ to the integration locus $\mathcal{I}$; freeze all of $\theta$
\State $\Delta^\star \gets \arg\min_{\Delta_{\mathcal{I}}} \mathbb{E}[-\log p_{\theta\oplus\Delta_{\mathcal{I}}}(\rho(y_q)\mid P_K)]$
\State \Return immunized model $f^{\,\mathrm{vac}}=f_{\theta\oplus\Delta^\star}$ \Comment{reused across all downstream tasks}
\Statex \textit{// Per-deployment serving with demonstration block $D=\{(x_i,y_i)\}$}
\State Estimate copy rate $\hat{r}(K)$ on a small held-out probe (Eq.~\eqref{eq:copyrate})
\State $K^\star \gets \max\{K : \hat{r}(K) \le \tau\}$ \Comment{gate: stop before saturation, Eq.~\eqref{eq:gate}}
\If{context budget is tight}
 \State $v \gets v_{\mathcal{I}}(D)$ task vector from $f^{\,\mathrm{vac}}$ at locus $\mathcal{I}$ (Eq.~\eqref{eq:inject})
 \State answer query with $v$ injected, spending no demonstration context \Comment{inject}
\Else
 \State answer query with the first $K^\star$ demonstrations in context
\EndIf
\end{algorithmic}
\end{algorithm}

\section{Empirical Validation}\label{sec:empirical}

We now validate empirically every claim previewed in Sec.~\ref{sec:proposed}. We first describe the
experimental setup, then report \textbf{(a)} the in-context collapse and where it lives, \textbf{(b)}
its repair by \textsc{CircA}, and \textbf{(c)} deeper analyses and ablations. All code, the
synthetic-concept generator, the intervention harness, and the result files behind every figure and
table are publicly available at \url{https://github.com/rostami-m/incontext-collapse}.

\subsection{Experimental setup}\label{sec:setup}

\subsubsection{Models and access}
Our open-weight panel comprises $11$ VLMs spanning three connector/backbone
families and a $0.5$--$11$B ($22\times$) scale range: MLP-projector VLMs
(Qwen2-VL-2B/7B \cite{wang2024qwen2vl}, Qwen2.5-VL-3B/7B, LLaVA-1.5-7B \cite{liu2023llava},
LLaVA-1.6-Mistral-7B, LLaVA-OneVision-0.5B/7B \cite{li2024llavaonevision}), pixel-shuffle VLMs
(SmolVLM-2B and SmolVLM-500M), and a cross-attention VLM (Llama-3.2-11B-Vision
\cite{meta2024llama32}),
chosen to decorrelate scale from connector design. The lesion-and-rescue cross-architecture replication
additionally uses InternVL3-2B \cite{chen2024internvl}. Behavioural sweeps were run on self-hosted checkpoints in
\texttt{bfloat16} with each model's native chat template applied explicitly, so that inference-time
prompts match the format the model was aligned on. The white-box analyses, the lesion-and-rescue, its
cross-architecture replication, and the continual-learning study, require access to hidden states and
weights and were run on the open checkpoints on single $24$\,GB GPUs, with $4$-bit adapters where memory
required and the quantization held fixed across the before/after comparison so it cannot affect a
difference score. For frontier scale we probe the Amazon Nova (Lite, Pro) and Anthropic Claude
(Haiku\,4.5, Sonnet\,4.5) inference APIs at default decoding. Q-Former models
\cite{li2023blip2,dai2023instructblip} are single-image architectures that cannot represent interleaved
multi-image prompts and are therefore out of scope for image--label ICL.

\subsubsection{Tasks and the synthetic-concept generator}
\begin{figure}[!ht]
\centering
\includegraphics[width=\textwidth]{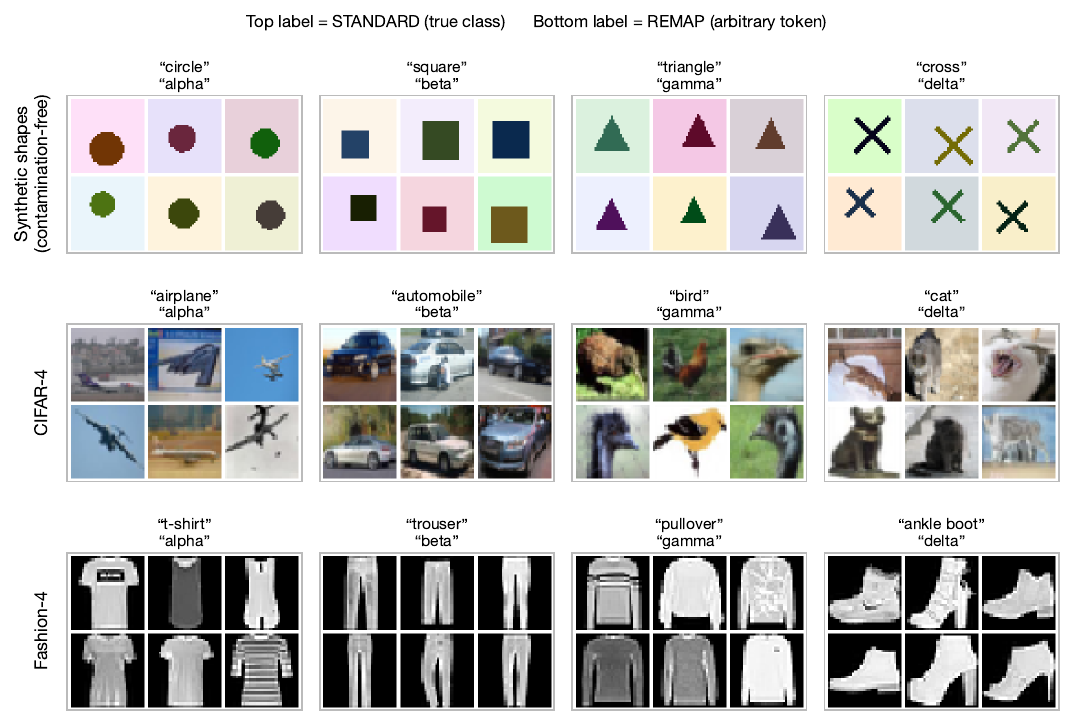}
\caption{\textbf{The task suite and the two label spaces.} One representative image per class for each
of the three $4$-way tasks: procedurally generated \emph{synthetic shapes} (contamination-free), CIFAR-4
(strong natural-image priors), and Fashion-4 (fine-grained). Each image is annotated with its
\textsc{standard} label (true class name, top) and its \textsc{remap} label (an arbitrary token such as
``alpha'' that cannot have appeared in pretraining, bottom). Above-chance \textsc{remap} accuracy can
therefore come only from learning the in-prompt image$\rightarrow$token rule, never from retrieval.}
\label{fig:tasks}
\end{figure}

The classification suite has three $4$-way tasks chosen to vary the strength of pretraining priors:
CIFAR-4 (strong natural-image priors), Fashion-4 (fine-grained), and a procedurally generated
\emph{synthetic shapes} task. Each shapes image contains one of four geometric primitives with
randomized colour, size and position; the shape$\rightarrow$label rule is assigned arbitrary tokens and
cannot have been seen in pretraining, so above-chance accuracy can only be produced by learning the
mapping in context. We note that synthetic data training is itself a source of degradation in
iterative VLM training loops \cite{hu2025multimodalcollapse}; our synthetic task is used only for
evaluation and adapter training, never for pretraining, so these failure modes are orthogonal. Demonstration pools and query sets are class-balanced and disjoint; chance accuracy
is $0.25$ throughout. We sweep the number of shots $K\in\{0,1,2,4,8,16\}$.
Figure~\ref{fig:tasks} shows representative stimuli from each task together with their \textsc{standard}
and \textsc{remap} labels.

\subsubsection{Conditions and the copy-rate diagnostic}
\begin{figure}[!ht]
\centering
\includegraphics[width=\textwidth]{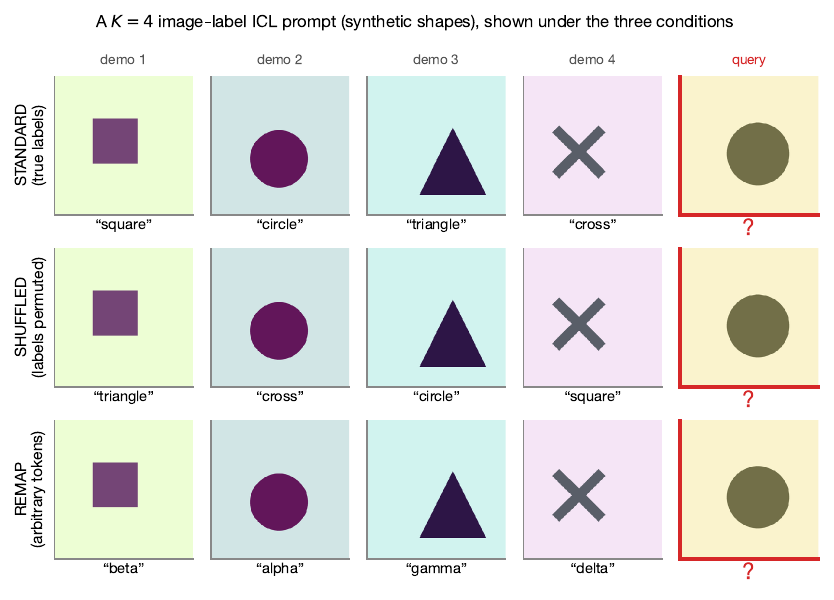}
\caption{\textbf{An image--label ICL prompt under the three conditions.} A $K{=}4$ prompt on synthetic
shapes: four (image, label) demonstrations followed by a held-out query (red). \textsc{standard} pairs
each image with its true label; \textsc{shuffled} keeps the same images but permutes the labels (so a
model riding its priors is unaffected, while a model integrating demonstration content is misled);
\textsc{remap} replaces the labels with arbitrary tokens (so only genuine in-context learning can solve
it). The model must answer the query (``?'') from the demonstrations alone.}
\label{fig:prompt}
\end{figure}

Each demonstration set is presented in three conditions. \textsc{standard} uses the true class names.
\textsc{shuffled} permutes the demonstration labels: if $\textsc{shuffled}\!\approx\!\textsc{standard}$
the model is ignoring demonstration content and riding its priors. \textsc{remap} assigns arbitrary
tokens to the classes, so only a model that learns the in-prompt mapping can exceed chance. As a
mechanism probe we additionally record the \emph{copy rate}: the fraction of predictions equal to the
label of the most recent demonstration. A high copy rate together with
$\textsc{shuffled}\!\approx\!\textsc{standard}$ identifies a model keying on demonstration
format/recency rather than integrating demonstration content. Figure~\ref{fig:prompt} renders a
four-shot prompt under all three conditions, making concrete what the model is shown in each.

\subsubsection{Prompt construction and decoding}\label{sec:setup-prompt}
For a task and shot count $K$, a prompt interleaves $K$ class-balanced (image, label) demonstrations
sampled without replacement with a single held-out query image, formatted with the model's chat
template; LLaVA-OneVision models use a multi-turn template, as their processor requires. The query is
never among the demonstrations, and images are presented at $112\times112$. Generation is greedy and
length-limited to the label span; outputs are mapped to the label set by constrained string matching,
and an output matching no label is recorded as unparseable. We evaluate $40$ held-out queries per cell
for the intervention study and $25$ for the panel, and report the unparseable fraction alongside
accuracy so that a formatting failure is not mistaken for a reasoning collapse.

A note on the shot range is in order. Each demonstration in a VLM prompt carries a full image, which a
modern vision encoder expands into hundreds to thousands of tokens; a single $K$ therefore costs far
more context than the same $K$ in a text-only prompt. At $K{=}64$ the longest prompts already exceed
$10^5$ tokens, SmolVLM cannot run that cell within its $16$k window at all, so our sweep up to
$K{=}64$ spans the same context-budget regime that ``many-shot'' denotes for text, even though the shot
count is numerically smaller. The collapse, moreover, sets in well before this ceiling, typically by
$K{=}4$--$8$ (Fig.~\ref{fig:failuremodes}), so it is a property of the early accumulation of
demonstrations rather than of an exotic long-context regime.

\subsubsection{Defining the collapse and classifying regimes}
For each model--task pair we obtain an accuracy-versus-$K$ curve under each condition. A pair is
\emph{collapse-prone} if \textsc{standard} accuracy at some $K>0$ falls below the zero-shot value by a
margin exceeding seed-to-seed variability, and \emph{collapse-resistant} otherwise. Orthogonally, a pair is a
\emph{learner} if \textsc{remap} accuracy rises above chance with $K$, and \emph{retrieval-only}
otherwise. The two binary axes define the three observed regimes; the full grid is in
Table~\ref{tab:apppanel}.

\subsubsection{Parameter-efficient lesion-and-rescue}\label{sec:setup-lesion}
This is the experimental backbone of our causal claim. The
intervention adds a small, region-restricted block of trainable capacity to an otherwise frozen model
and asks whether that capacity, placed at a particular depth, restores in-context learning. ``Lesion''
refers to the fact that the unintervened model is functionally deficient at the integration step;
``rescue'' to adding capacity exactly where we hypothesize the deficit lives. Because the model is frozen
everywhere else, any recovery is attributable to the adapter's \emph{location}.

\paragraph{Adapter and optimization.}
The trainable block is a LoRA adapter \cite{hu2022lora}: for each targeted linear weight
$W\in\mathbb{R}^{d_{\mathrm{out}}\times d_{\mathrm{in}}}$ we learn a low-rank update
$W \mapsto W + \tfrac{\alpha}{r}BA$ with $B\in\mathbb{R}^{d_{\mathrm{out}}\times r}$,
$A\in\mathbb{R}^{r\times d_{\mathrm{in}}}$, rank $r{=}8$, scaling $\alpha{=}16$ (so $\alpha/r{=}2$), and
LoRA dropout $0.05$; $A$ is Gaussian-initialized and $B$ is zero-initialized, so the adapted model is
identical to the base model at step $0$ and the intervention can only \emph{add} capability. We optimize
the adapter with AdamW at learning rate $10^{-4}$ for $150$ steps (batch size $1$), with all base
parameters $\theta$ frozen; gradient checkpointing and ``use\_cache$=$false'' keep activation memory
bounded so the whole study fits on a single $24$\,GB GPU. 


\paragraph{Regions and the equal-capacity control.}
A \emph{region} is a set of linear modules selected by depth. Following the partition of
Eq.~\eqref{eq:partition}, we define five regions: the \emph{connector} (the projection/resampler linear
modules bridging the vision encoder to the language model), the \emph{early}/\emph{mid}/\emph{late}
thirds of the language-model blocks, and \emph{all} blocks. Each language-model region is additionally
split into attention-only ($q,k,v,o$ projections) and MLP-only (gate/up/down projections) variants, for
the component cut of Sec.~\ref{sec:setup-kspec}. The number of adapter modules per region is capped (at
$64$ language-model modules, $8$ connector modules) so that the compared regions carry \emph{matched}
parameter budgets: on Qwen2-VL-2B the early and late regions differ by under $10\%$ of modules ($70$
vs.\ $63$). This equal-capacity control is essential, because a region could otherwise appear to rescue
simply by having more parameters; with budget held fixed, a difference between rescuing at the
integration locus and at the readout reflects \emph{location}, not capacity. \emph{Targeting is verified
at run time}: after construction we enumerate the instantiated adapter modules and assert every one falls
within the intended region, so a region label always denotes exactly the modules it names.

\paragraph{What is trained, and why success is meaningful.}
Each task supplies a class-balanced training pool and a held-out evaluation set; we further split the
evaluation items so the adapter is trained and measured on \emph{disjoint} queries (no query seen in
training is scored). At each step we sample a fresh $K{=}8$-shot demonstration block and a query, format
them with the same interleaved prompt used everywhere (Sec.~\ref{sec:setup-prompt}), and apply the
crucial supervision asymmetry: the in-context demonstrations carry the \emph{arbitrary} \textsc{remap}
labels $\rho$, while the loss supervises the query's \emph{true} class name, with the loss masked to the
answer tokens only. The adapter therefore cannot succeed by memorizing any fixed label$\rightarrow$answer
association, the remapped labels make that useless; it can only succeed by learning to \emph{read the
demonstrations and apply their mapping to the query}. After training we re-measure the full standard and
\textsc{remap} $K$-curves by greedy generation and the same answer parser as the panel. Success, a
\textsc{remap} curve that now rises with $K$ where the frozen model's was flat at chance, thus certifies
that the adapter restored the \emph{ability to use demonstrations}, the exact capability the collapse
removes. The complementary \emph{induce} arm inverts the supervision on a robust learner (demonstrations
carry a conflicting shuffled mapping, supervision targets the prior) to test whether the collapse can be
\emph{installed} at the same locus.

\subsubsection{K-specificity, component cut and cross-architecture replication}\label{sec:setup-kspec}
K-specificity is read from the full rescue curve (Table~\ref{tab:appcurves}): the integration--late gap
is negligible at $K{\le}2$ and maximal at $K{=}8,16$. The component cut restricts the within-region
adapter to the attention projections or to the MLP projections only. Cross-architecture replication
repeats the full region protocol on InternVL3-2B; because its integration capacity is higher, we use a
larger adapter (rank $32$, $400$ steps), and we report it as a directional replication.

\subsubsection{Frontier probe}
Frontier models are evaluated through the Converse API with interleaved image and text content blocks,
on CIFAR-4 and synthetic shapes, sweeping $K\in\{0,1,2,4,8\}$ over $24$--$40$ balanced queries. The
\textsc{remap} learning curves are reported as the mean over three seeds (with $95\%$ CIs); the
\textsc{standard} collapse curves are single-seed probes over the full balanced query set.

\subsubsection{Continual learning protocol}
The continual-learning stream splits CIFAR-100 into five disjoint $4$-class tasks presented
sequentially. After each task we fit a LoRA restricted to one of three loci, the circuit
(connector\,+\,early/mid), all layers, or late layers, and evaluate every task seen so far. Average
accuracy (AA) is the mean over tasks after the final task; backward transfer (BWT) is the mean change in
each task's accuracy from when it was learned to the end of the stream. We report mean $\pm$ s.d.\ over
three seeds.

\subsubsection{Statistics, compute and reproducibility}
Intervention and continual-learning results are reported as mean $\pm$ $95\%$ confidence interval
(intervention) or $\pm$ s.d.\ (continual learning) over three seeds; panel sweeps use two seeds. No runs
were excluded. All white-box cells complete in under an hour on a single GPU, so the full localization,
component-cut, cross-architecture and continual-learning panels are inexpensive to reproduce. All prompt
templates, the synthetic-concept generator, the intervention harness, the figure-generation script, and
the per-seed result files are released.


\subsection{The in-context collapse: prevalence, regimes, and where it lives}\label{sec:resultsA}

We first establish that the collapse is real and graded, that it is dissociable from genuine
in-context learning, that it reaches standard VQA benchmarks and frontier scale, and that it is
\emph{causally localized} to the vision--language integration pathway.

\subsubsection{A graded in-context collapse across the panel}\label{sec:cliff}
\begin{figure}[!ht]
\centering
\includegraphics[width=\textwidth]{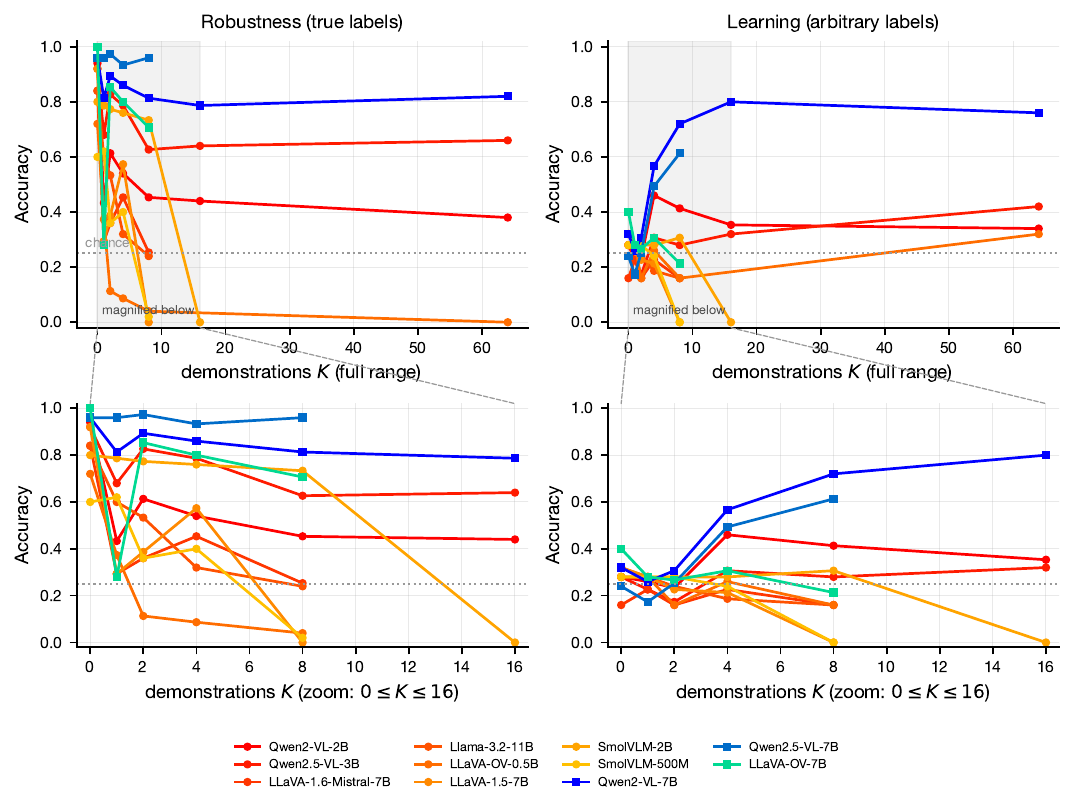}
\caption{\textbf{The in-context collapse and the robustness--learning dissociation, across the panel.}
All eleven open VLMs in the panel, on CIFAR-4. \textbf{(a)}~\textsc{standard} accuracy versus the
number of demonstrations $K$. Collapse-prone models (warm) lose accuracy as demonstrations accumulate,
from mild (Qwen2.5-VL-3B) to catastrophic (LLaVA/SmolVLM families fall to or below chance), while
collapse-resistant models (cool) remain stable. The effect spans MLP-projector, cross-attention, and
high-token architectures and does not track parameter count. \textbf{(b)}~On \textsc{remap} (arbitrary
labels; chance $=0.25$, dotted) only the well-aligned large models (Qwen2-VL-7B, Qwen2.5-VL-7B) rise
with $K$, i.e.\ genuinely \emph{learn} the in-prompt rule; the prone models stay at chance,
\emph{retrieving} priors rather than learning. Robustness (panel a) and learning (panel b) are thus
separable axes (Table~\ref{tab:apppanel}). \textbf{Top row:} the full shot range (to $K{=}64$);
the shaded band is magnified in the \textbf{bottom row} ($0\!\leq\!K\!\leq\!16$), where most of the
dynamics, the first-shot drop and the slow rise of the learners, occur.}
\label{fig:cliff}
\end{figure}

We evaluate every model in the panel on all three classification tasks (synthetic shapes, CIFAR-4,
Fashion-4) under both verbalizers, and summarize the complete grid in Table~\ref{tab:apppanel}. The
headline is that the collapse is real and \emph{graded}: on \textsc{standard} classification many VLMs
are strong zero-shot classifiers whose accuracy \emph{falls} as demonstrations are added
(Fig.~\ref{fig:cliff}a), but how far it falls varies sharply from model to model. We measure $11$ open
VLMs spanning $0.5$--$11$B parameters and several connector designs (MLP projectors, pixel-shuffle,
cross-attention, perceiver/resampler), and the effect ranges from negligible to catastrophic.

\paragraph{The phenomenon is broad, not a quirk of one model or one task.}
On CIFAR-4 the collapse appears, to differing degrees, in eight of the eleven models. It is mild in some
(Qwen2.5-VL-3B, $0.92\!\rightarrow\!0.64$ by $K{=}16$), substantial in many (Qwen2-VL-2B
$0.94\!\rightarrow\!0.44$; LLaVA-1.6-Mistral-7B $0.84\!\rightarrow\!0.25$; Llama-3.2-11B
$0.84\!\rightarrow\!0.24$), and catastrophic in the smallest or least-aligned (LLaVA-OV-0.5B
$0.72\!\rightarrow\!0.04$ by $K{=}8$; LLaVA-1.5-7B and SmolVLM-500M fall to chance and then degenerate).
A separate group is essentially flat (Qwen2-VL-7B $0.96\!\rightarrow\!0.79$; Qwen2.5-VL-7B
$0.96\!\rightarrow\!0.96$; LLaVA-OV-7B holds the high $0.7$s). The same pattern reproduces
across all three tasks (Table~\ref{tab:apppanel}): a given model's regime is largely stable across
shapes, CIFAR-4, and Fashion-4, with the contamination-free shapes task tending to give the highest
zero-shot accuracy (no domain gap) and therefore the steepest visible collapse. Crucially, the collapse
spans \emph{architecture families}, it is present in MLP-projector (Qwen, LLaVA-1.5/1.6), cross-attention
(Llama-3.2-11B), and high-token (LLaVA-OV) designs alike, so it is a property of how a VLM is aligned to
follow many-shot prompts rather than of any single connector design.

\paragraph{Susceptibility tracks the alignment recipe, not parameter count.}
Scale alone does not predict the regime. An $11$B cross-attention model (Llama-3.2-11B) and a $7$B
MLP-projector model (LLaVA-1.6-Mistral) both collapse substantially, while a $7$B model from a different
recipe (Qwen2-VL-7B) and even a $0.5$--$2$B SmolVLM-2B do not; conversely within one model family the
$2$B Qwen collapses where the $7$B does not. What separates the regimes is consistent: the
collapse-resistant models are those whose post-training makes them treat demonstrations as evidence to
be integrated, the collapse-prone models revert to a recency heuristic (quantified in
Sec.~\ref{sec:dissoc}). \emph{Takeaway: many-shot multimodal ICL is not uniformly safe; whether it helps
or harms is set by the alignment recipe, and a practitioner cannot infer immunity from model size.}

Beyond prevalence, three properties of the curves sharpen what the collapse is and rule out innocent
explanations; we illustrate them on the clean reasoning case (Qwen2-VL-2B) and verify they generalize
across the prone models.

\begin{table}[!ht]
\centering
\caption{\textbf{The full panel: robustness and learning by model and task} ($11$ open VLMs, mean over
$3$ seeds; SmolVLM-500M over $2$). $\mathrm{std}$ is \textsc{standard} accuracy (robustness),
$\mathrm{rmp}$ is \textsc{remap} accuracy (learning; chance $=0.25$), each as
$K{=}0\!\rightarrow\!K{=}\max$, with $\max{=}16$ for Qwen2-VL-2B/7B, $8$ for most others, and $4$ for
the single-image models LLaVA-1.5-7B and SmolVLM-500M (whose outputs degenerate beyond that point).
Falling $\mathrm{std}$ indicates the collapse; rising $\mathrm{rmp}$ indicates genuine learning. Models
are grouped into the three regimes of Sec.~\ref{sec:dissoc}.}
\label{tab:apppanel}
\begin{tabular}{llcccc}
\toprule
Regime & Model & Task & $\mathrm{std}$ ($0\!\rightarrow\!\max$) & $\mathrm{rmp}$ ($0\!\rightarrow\!\max$) \\
\midrule
\multirow{17}{*}{\shortstack[l]{(i) prone,\\ retrieval}}
 & \multirow{3}{*}{Qwen2-VL-2B} & shapes & $1.00\!\rightarrow\!0.44$ & $0.24\!\rightarrow\!0.36$ \\
 & & cifar & $0.94\!\rightarrow\!0.44$ & $0.28\!\rightarrow\!0.35$ \\
 & & fashion & $0.88\!\rightarrow\!0.35$ & $0.34\!\rightarrow\!0.21$ \\
 & \multirow{3}{*}{Qwen2.5-VL-3B} & shapes & $1.00\!\rightarrow\!0.80$ & $0.44\!\rightarrow\!0.35$ \\
 & & cifar & $0.92\!\rightarrow\!0.64$ & $0.28\!\rightarrow\!0.32$ \\
 & & fashion & $0.84\!\rightarrow\!0.28$ & $0.16\!\rightarrow\!0.25$ \\
 & \multirow{3}{*}{LLaVA-1.6-Mistral-7B} & shapes & $1.00\!\rightarrow\!0.40$ & $0.32\!\rightarrow\!0.23$ \\
 & & cifar & $0.84\!\rightarrow\!0.25$ & $0.16\!\rightarrow\!0.16$ \\
 & & fashion & $0.96\!\rightarrow\!0.21$ & $0.36\!\rightarrow\!0.37$ \\
 & \multirow{3}{*}{Llama-3.2-11B} & shapes & $1.00\!\rightarrow\!0.45$ & $0.08\!\rightarrow\!0.24$ \\
 & & cifar & $0.84\!\rightarrow\!0.24$ & $0.28\!\rightarrow\!0.16$ \\
 & & fashion & $0.96\!\rightarrow\!0.41$ & $0.28\!\rightarrow\!0.24$ \\
 & \multirow{3}{*}{LLaVA-OV-0.5B} & shapes & $1.00\!\rightarrow\!0.28$ & $0.20\!\rightarrow\!0.34$ \\
 & & cifar & $0.72\!\rightarrow\!0.04$ & $0.32\!\rightarrow\!0.16$ \\
 & & fashion & $0.16\!\rightarrow\!0.16$ & $0.16\!\rightarrow\!0.30$ \\
 & LLaVA-1.5-7B & cifar & $0.92\!\rightarrow\!0.57$ & $0.32\!\rightarrow\!0.21$ \\
 & SmolVLM-500M & cifar & $0.60\!\rightarrow\!0.40$ & $0.28\!\rightarrow\!0.24$ \\
\midrule
\multirow{4}{*}{\shortstack[l]{(ii) immune,\\ retrieval}}
 & \multirow{3}{*}{SmolVLM-2B} & shapes & $1.00\!\rightarrow\!1.00$ & $0.28\!\rightarrow\!0.32$ \\
 & & cifar & $0.80\!\rightarrow\!0.73$ & $0.28\!\rightarrow\!0.31$ \\
 & & fashion & $0.76\!\rightarrow\!0.68$ & $0.16\!\rightarrow\!0.13$ \\
 & LLaVA-OV-7B & cifar & $1.00\!\rightarrow\!0.71$ & $0.40\!\rightarrow\!0.21$ \\
\midrule
\multirow{6}{*}{\shortstack[l]{(iii) immune,\\ learner}}
 & \multirow{3}{*}{Qwen2-VL-7B} & shapes & $1.00\!\rightarrow\!0.99$ & $0.50\!\rightarrow\!0.98$ \\
 & & cifar & $0.96\!\rightarrow\!0.79$ & $0.32\!\rightarrow\!0.80$ \\
 & & fashion & $0.88\!\rightarrow\!0.77$ & $0.32\!\rightarrow\!0.56$ \\
 & \multirow{3}{*}{Qwen2.5-VL-7B} & shapes & $1.00\!\rightarrow\!1.00$ & $0.48\!\rightarrow\!1.00$ \\
 & & cifar & $0.96\!\rightarrow\!0.96$ & $0.24\!\rightarrow\!0.61$ \\
 & & fashion & $0.92\!\rightarrow\!0.97$ & $0.16\!\rightarrow\!0.83$ \\
\bottomrule
\end{tabular}
\end{table}

\begin{figure}[!ht]
\centering
\includegraphics[width=\textwidth]{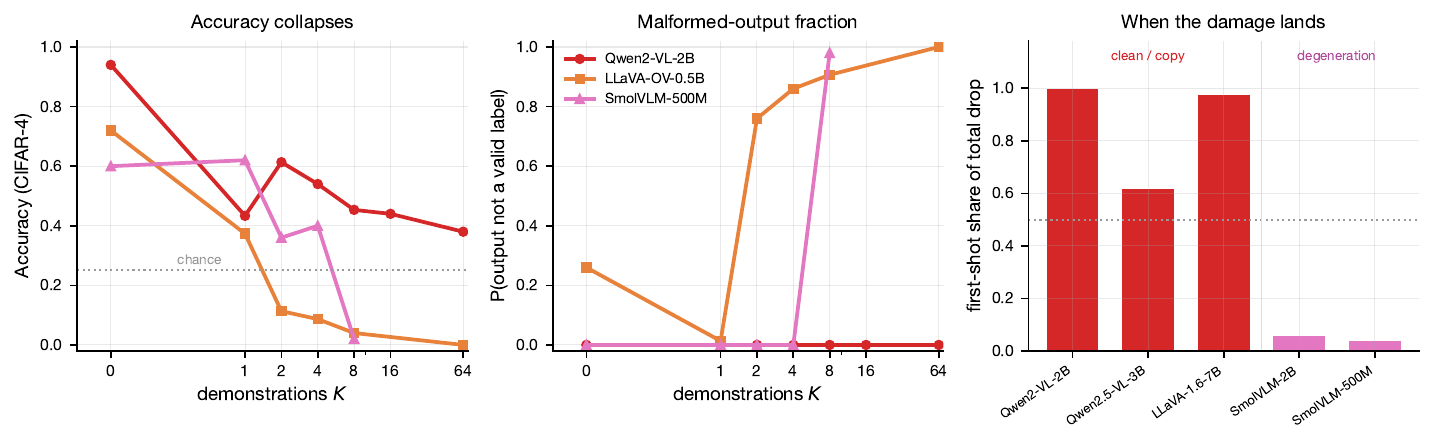}
\caption{\textbf{Two failure modes with distinct signatures; the collapse is not a formatting artifact.}
\textbf{(left)}~\textsc{standard} accuracy on CIFAR-4 versus $K$ (log-spaced) for three collapse-prone
models. \textbf{(middle)}~The fraction of outputs that are not a valid label. Qwen2-VL-2B (red) loses
half its accuracy with a malformed-output fraction pinned at \emph{zero}, a clean reasoning collapse.
The below-chance models (LLaVA-OV-0.5B, SmolVLM-500M) instead couple their accuracy crash to a surge in
malformed outputs, so their sub-chance scores partly reflect output degeneration.
\textbf{(right)}~The two modes also differ in \emph{timing}: the share of the total
\textsc{standard}-accuracy drop that lands on the very first demonstration ($K{=}0\!\rightarrow\!1$),
averaged over the tasks each model collapses on. The clean-reasoning/copy models (red) front-load
$86$--$100\%$ of the damage onto the first shot, whereas the degeneration models (pink) lose almost
nothing there and erode gradually as their outputs fall apart with context length. Mechanistic claims in
this paper are anchored on the clean case.}
\label{fig:failuremodes}
\end{figure}

\paragraph{The damage is done at the very first demonstration, in the clean-reasoning collapse.}
In the models whose collapse is a genuine reasoning failure (zero malformed outputs;
Fig.~\ref{fig:failuremodes}), the collapse is not a slow erosion that accumulates with context length;
it is almost a step change at $K{=}1$. For Qwen2-VL-2B on CIFAR-4 the single largest move in the entire
curve is the $K{=}0\!\rightarrow\!K{=}1$ drop of $0.51$ ($0.94\!\rightarrow\!0.43$), which already
accounts for the overwhelming majority of the total $0.56$ decline out to $K{=}64$; no later change
between adjacent shot counts exceeds $0.18$, and the curve only drifts downward from there. This
front-loading is the rule across the clean-reasoning models: the first demonstration alone delivers
$86$--$100\%$ of the entire \textsc{standard}-accuracy drop, averaged over the three tasks, for
Qwen2-VL-2B, Qwen2.5-VL-3B, and LLaVA-1.6-Mistral-7B alike. Strikingly, the \emph{degeneration}-mode
models behave oppositely: SmolVLM-2B/500M, whose accuracy crash is coupled to a surge in malformed
outputs, lose almost nothing on the first shot ($\le2\%$ of their total drop) and instead erode
gradually as their outputs progressively fall apart with context length. \emph{Takeaway: the
clean-reasoning collapse is triggered by the \emph{presence} of demonstrations, not by their accumulation
past some context budget, a single in-context example is enough to knock a confident zero-shot classifier
off its prior, and the timing itself separates the two failure modes: a copy-driven reasoning collapse
strikes instantly, whereas output degeneration accrues with length.} The first observation already argues
against a long-context or token-budget explanation for the clean case, since the effect is
saturated long before context length is a concern (we return to this with the $K{=}64$ runs below).

\begin{figure}[!ht]
\centering
\includegraphics[width=\textwidth]{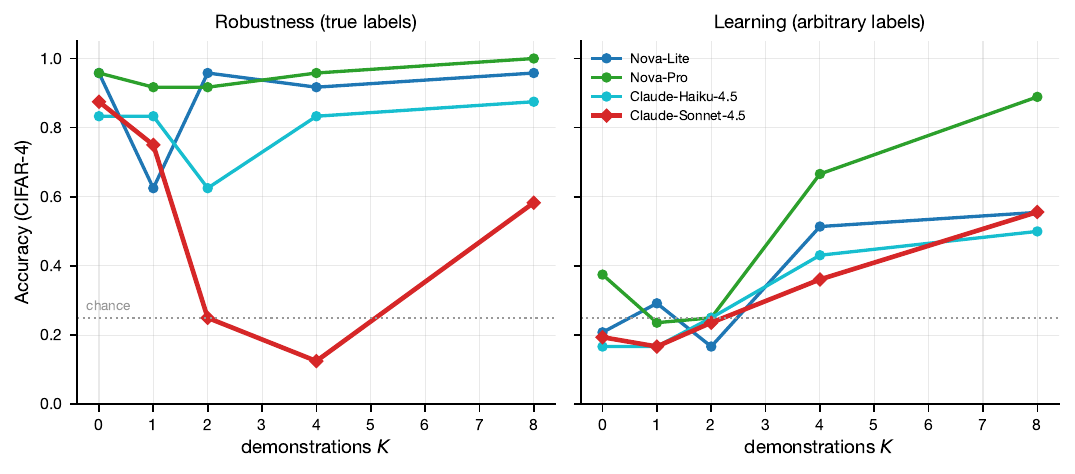}
\caption{\textbf{The collapse, and the robustness--learning dissociation, reach frontier scale.}
Four frontier models (inference APIs) on CIFAR-4. \textbf{(a)}~\textsc{standard} accuracy versus $K$:
Nova-Lite/Pro and Claude-Haiku-4.5 are robust (cool), while Claude\,Sonnet\,4.5 (red) collapses from
$0.88$ to $0.12$ as a handful of demonstrations are added, then only partially recovers. \textbf{(b)}~On
\textsc{remap} (arbitrary labels; chance $=0.25$, dotted) \emph{all four}, including the collapse-prone
Sonnet\,4.5, \emph{rise} with $K$, i.e.\ genuinely learn the in-prompt rule (to $0.50$--$0.89$ at
$K{=}8$). Thus a frontier model can collapse on natural labels yet still learn an arbitrary mapping: the
two axes dissociate even at frontier scale, exactly as in the open panel (Fig.~\ref{fig:cliff}). The
collapse is recipe/model-specific, not a small-model artifact.}
\label{fig:frontier}
\end{figure}

\paragraph{The performance curves are non-monotonic, and they do not recover.}
Several collapse-prone models trace a shallow V: Qwen2-VL-2B dips to $0.43$ at $K{=}1$, partially
rebounds to $0.61$ at $K{=}2$, then settles in the low-$0.4$s; Qwen2.5-VL-3B and, at frontier scale,
Claude\,Sonnet\,4.5 (Fig.~\ref{fig:frontier}) show the same dip-and-partial-recovery. This matches the
``V-shaped'' many-shot curves reported on closed models \cite{jiang2024manyshotmm}, but with a crucial
difference: in our collapse-prone models accuracy never climbs back to the zero-shot value, even at
$K{=}64$. \emph{Takeaway: the collapse is not a transient few-shot artifact that more demonstrations
cure; the prior, once displaced, is not recovered by piling on examples.}

\paragraph{The collapse does not wash out at large $K$.}
Pushing to $K{=}64$ (where prompts exceed $10^5$ visual tokens) does not reverse the collapse: Qwen2-VL-2B
sits at $0.38$ and LLaVA-OV-0.5B at $0.00$. \emph{Takeaway: this is not a small-context phenomenon that
many-shot prompting would outgrow; it persists, and in the weakest models deepens, across two orders of
magnitude in shot count.}

\begin{figure}[!ht]
\centering
\includegraphics[width=\textwidth]{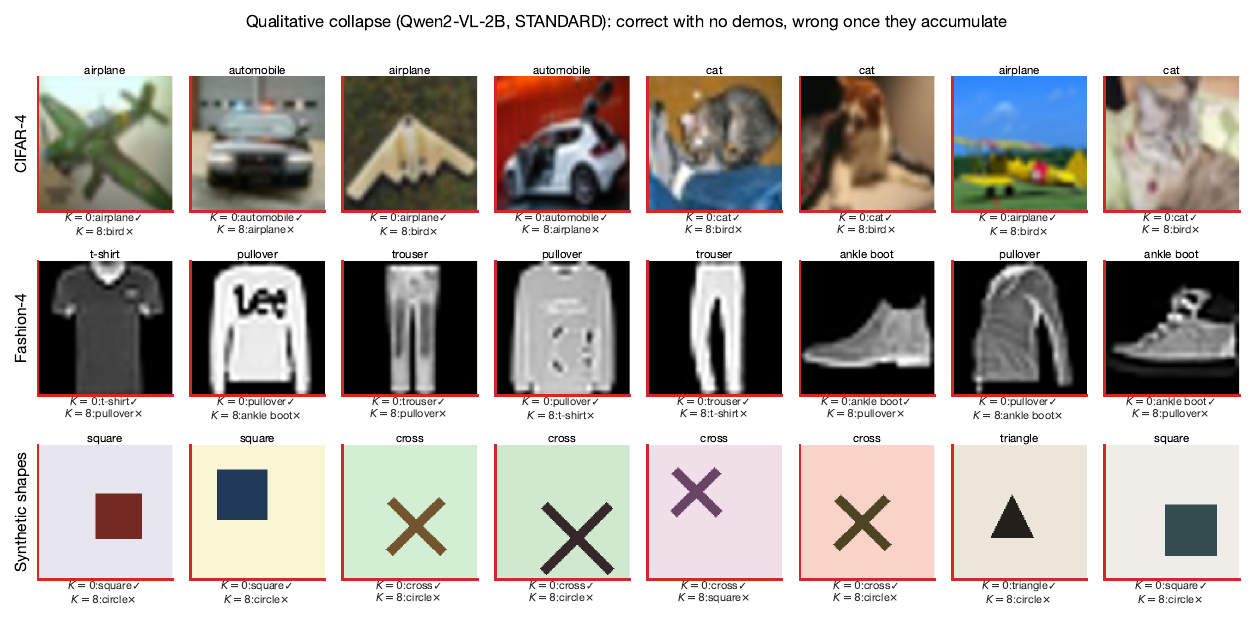}
\caption{\textbf{The collapse, query by query (Qwen2-VL-2B, \textsc{standard}).} For eight held-out
images per task, the model's prediction with no demonstrations ($K{=}0$) and after eight correctly
labelled demonstrations ($K{=}8$). With no demonstrations every image is classified correctly; adding
demonstrations flips it to a \emph{different in-task class} (e.g.\ airplane$\rightarrow$bird,
cat$\rightarrow$bird, t-shirt$\rightarrow$pullover, square$\rightarrow$circle). The outputs remain
well-formed labels throughout, the failure is one of judgement, not formatting. Green/red borders mark
the $K{=}8$ outcome.}
\label{fig:gallery}
\end{figure}

\paragraph{Two failure modes, and why this is not a formatting artifact.}
A natural worry is that the collapse is merely a degradation of \emph{output format}, the model
ceasing to emit a parseable label, rather than a degradation of the underlying decision. The data
distinguish two modes (Fig.~\ref{fig:failuremodes}). In the first, the model continues to emit
well-formed labels but chooses the wrong one: Qwen2-VL-2B's malformed-output fraction stays at
\emph{zero} across all $K$ even as its accuracy falls by half, so its collapse is a genuine reasoning
failure that no relaxation of output parsing could explain. In the second, the below-chance cases
(LLaVA-OneVision-0.5B, SmolVLM-500M) couple their accuracy crash to a surge in malformed outputs (to
$90$--$98\%$ by the collapse point), so part of their sub-chance accuracy is output degeneration under
demonstration overload. We therefore anchor every mechanistic and causal claim in this paper on the
clean reasoning collapse (Qwen2-VL-2B), and read the below-chance models as the extreme tail of the
same phenomenon rather than as its centre. Because the collapse appears in-distribution, demonstrations
and queries are drawn from the same task and label set, it cannot be attributed to the
support-set/query mismatch that drives degradation elsewhere \cite{huang2025rethinkmmicl}.
Figure~\ref{fig:gallery} makes the clean reasoning collapse tangible at the level of individual queries:
the model names each image correctly with no demonstrations, then mislabels it, usually as another
in-task class, once a handful of correctly-labelled demonstrations are prepended.


\paragraph{The collapse, and the dissociation, reach the frontier.}
The collapse is not an artifact of small open models (Fig.~\ref{fig:frontier}). Of four frontier models
probed on CIFAR-4, three are robust, but Claude\,Sonnet\,4.5 exhibits a sharp
collapse on \textsc{standard} labels, accuracy $0.88$ at $K{=}0$ falling to $0.12$ at $K{=}4$ before a
partial recovery, while remaining perfect on the contamination-free shapes task.
Strikingly, the robustness--learning dissociation (Sec.~\ref{sec:dissoc}) holds here too: on \textsc{remap}
all four models, \emph{including} the collapse-prone Sonnet\,4.5, \emph{rise} with $K$ and learn the
arbitrary mapping ($0.50$--$0.89$ at $K{=}8$).\footnote{The instruction-tuned frontier models answer
verbosely; for the \textsc{remap} probe we use a stricter single-word prompt so the (arbitrary) label is
emitted rather than a description. The prompt leaves the already-compliant models' scores essentially
unchanged.} That a frontier model collapses on natural labels yet still acquires a novel in-prompt rule
shows the collapse is a failure of \emph{robustness to prior-conflicting demonstrations}, not of in-context
learning per se, and that susceptibility is set by the model and its recipe, not by scale alone.

\subsubsection{Robustness and in-context learning are dissociable}\label{sec:dissoc}
The collapse measures only \emph{robustness}, whether accuracy survives added demonstrations, and says
nothing about whether the model can \emph{learn} from them. We probe learning with \textsc{remap}:
arbitrary class tokens that cannot be retrieved from priors, so above-chance accuracy requires
acquiring the in-prompt mapping. The two axes come apart (Fig.~\ref{fig:cliff}b), and the full panel
populates all three resulting regimes (Table~\ref{tab:apppanel}). A model can be
\emph{collapse-resistant yet unable to learn} (SmolVLM-2B, LLaVA-OV-7B: flat \textsc{standard} but
\textsc{remap} pinned near chance); \emph{collapse-prone} (Qwen2-VL-2B, LLaVA-1.6-Mistral-7B,
Llama-3.2-11B, LLaVA-OV-0.5B: \textsc{standard} collapses, \textsc{remap} at chance); or \emph{both
robust and a genuine learner} (Qwen2-VL-7B, Qwen2.5-VL-7B: \textsc{standard} high \emph{and}
\textsc{remap} rising, Qwen2-VL-7B reaches $0.80$ on CIFAR-4 and $0.98$ on shapes at $K{=}16$). These
combinations define three reproducible regimes that are populated across connector families
(MLP-projector, pixel-shuffle, cross-attention) and the full $0.5$--$11$B scale range, so regime
membership tracks the alignment recipe more than raw capacity.

\paragraph{Learning has the opposite temporal signature to the collapse.}
The two axes do not merely differ in their endpoints; they evolve in \emph{opposite} ways as
demonstrations accumulate, which is itself evidence that they are distinct computations. The collapse is
front-loaded, almost all of it lands on the first demonstration (above). Genuine learning is the mirror
image: it is absent at low $K$ and builds slowly. The one clear learner, Qwen2-VL-7B, shows essentially
no \textsc{remap} gain through $K{=}2$ ($0.32\!\rightarrow\!0.31$ on CIFAR-4) and then climbs steadily as
evidence accrues, $0.57$ at $K{=}4$, $0.72$ at $K{=}8$, $0.80$ at $K{=}16$; on the contamination-free
shapes task it reaches near-ceiling ($0.50\!\rightarrow\!0.99$). \emph{Takeaway: a collapse-prone model
breaks the instant a demonstration appears, whereas a learner needs several demonstrations before the
in-prompt rule pays off, so the very demonstrations that destroy one model's prior are what a learner is
only beginning to exploit.}

\paragraph{The collapse-prone model cannot be coaxed into learning by adding shots.}
Crucially, the prone model's failure on \textsc{remap} is not a slow start that more examples would fix:
Qwen2-VL-2B's shapes-\textsc{remap} accuracy is stuck near chance at every shot count
($0.24\!\rightarrow\!0.36$ across $K{=}0\ldots16$), never approaching the learner's trajectory. \emph{Takeaway:
no amount of in-context evidence lets a collapse-prone model acquire a novel image$\rightarrow$label
rule; the capability is absent, not merely slow, which is exactly what the causal intervention of
Sec.~\ref{sec:circuit} later restores.}

\begin{figure}[!ht]
\centering
\includegraphics[width=0.9\textwidth]{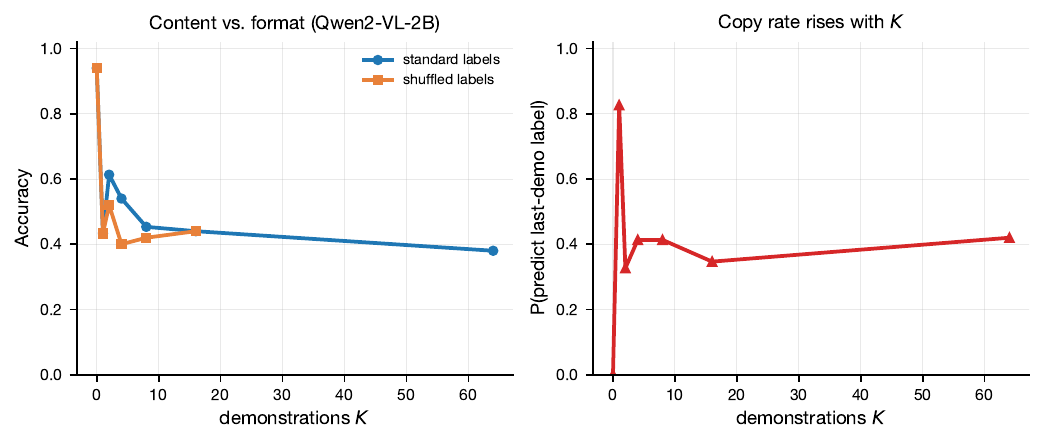}
\caption{\textbf{Content versus format (Qwen2-VL-2B, CIFAR-4).} \textbf{(left)}~\textsc{standard} and
\textsc{shuffled} accuracy track each other closely, so the model is largely insensitive to whether
demonstration labels are correct, it rides priors rather than integrating content.
\textbf{(right)}~The copy rate (fraction of predictions equal to the most recent demonstration's label)
rises with $K$, the signature of recency-driven copying rather than rule learning.}
\label{fig:control}
\end{figure}

\begin{figure}[!ht]
\centering
\includegraphics[width=\textwidth]{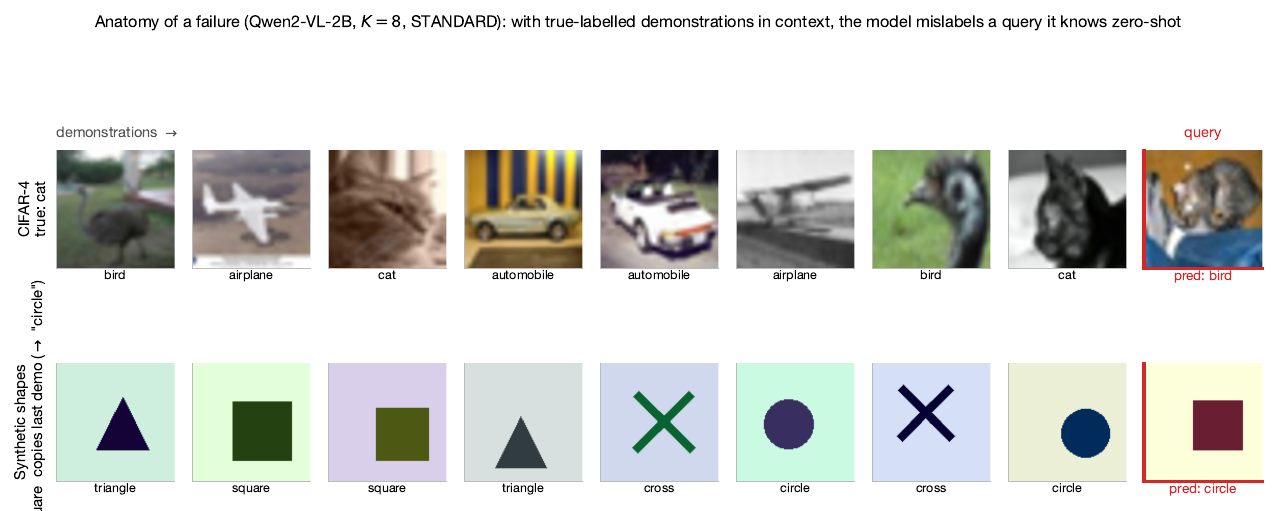}
\caption{\textbf{Anatomy of a failure (Qwen2-VL-2B, $K{=}8$, \textsc{standard}).} The complete eight
(image, true-label) demonstrations and the held-out query (red), for one CIFAR-4 and one synthetic-shapes
case. Both queries are classified correctly with no demonstrations; with the demonstrations in context
the model mislabels them, and on shapes it copies the most recent demonstration's label (``circle'')
instead of reading the query square. The outputs are valid labels, so this is a reasoning failure, not a
parsing one.}
\label{fig:failcase}
\end{figure}

\paragraph{Where the accuracy goes: copying the most recent label.}
The \textsc{shuffled} control confirms the reading: for regime-(i)/(ii) models
$\textsc{shuffled}\!\approx\!\textsc{standard}$ with high copy rates, i.e.\ they ignore demonstration
content; only regime-(iii) models are content-sensitive. The mechanism is visible in a tight coupling
between accuracy and the copy rate (Fig.~\ref{fig:control}). For Qwen2-VL-2B on CIFAR-4 the copy rate
jumps from $0$ at $K{=}0$ to $0.83$ at $K{=}1$, exactly the shot count at which accuracy craters to
$0.43$; thereafter both partially relax in lock-step (copy rate $\sim\!0.35$--$0.42$, accuracy
$\sim\!0.44$--$0.61$). In other words, the lost accuracy does not scatter into random errors, it is
captured almost entirely by predictions that echo the label of the most recent demonstration. \emph{Takeaway:
the collapse is a specific, legible failure, the model substitutes ``answer like the last example I was
shown'' for ``read the query'', and the recency-copy is strongest precisely at $K{=}1$ where the collapse
is steepest.} Figure~\ref{fig:failcase} shows this happening on a single prompt: an unmistakable query
square is labelled ``circle'', the label attached to the most recent demonstration.
The dissociation is also a practical diagnostic, it tells a practitioner which models can be adapted by
prompting alone, and it sharpens the mechanistic question: what, inside a collapse-prone model, prevents
demonstrations from being integrated?

\subsubsection{The collapse on open-ended VQA benchmarks}\label{sec:vqa}
\begin{figure}[!ht]
\centering
\includegraphics[width=\textwidth]{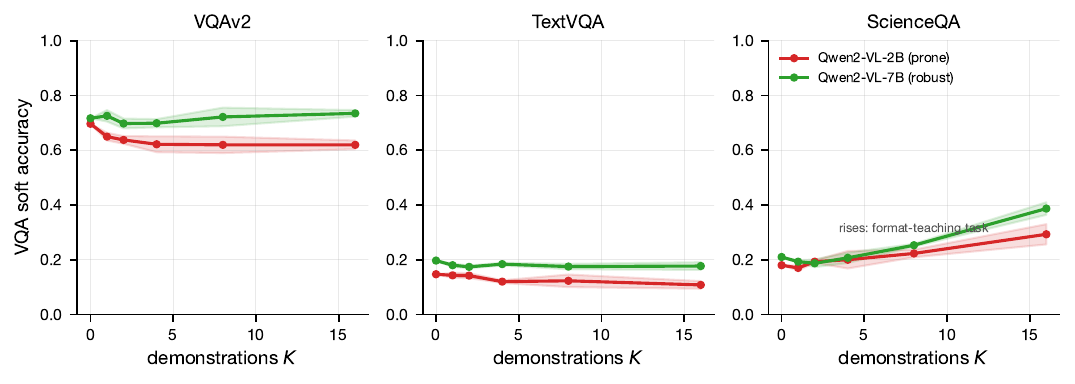}
\caption{\textbf{The collapse on open-ended VQA benchmarks.} VQA soft accuracy versus $K$ (mean
$\pm 95\%$ CI over $3$ seeds) for the collapse-prone Qwen2-VL-2B (red) and the robust Qwen2-VL-7B
(green). On the perceptual benchmarks (VQAv2, TextVQA) the $2$B model degrades while the $7$B holds,
the same ordering as on classification. On ScienceQA both \emph{rise}: its multiple-choice
demonstrations teach an answer \emph{format} rather than fight a perceptual prior, exactly the regime in
which demonstrations help.}
\label{fig:vqa_curves}
\end{figure}

Sec.~\ref{sec:cliff} established the collapse across three classification tasks and eleven models; a
remaining objection is that it is an artifact of the closed $4$-way classification framing itself. This
section is a targeted \emph{framing-robustness control}, and for a control the right design is not a wide
panel but a clean comparison that varies only the framing. We therefore use a within-family
\emph{controlled pair}: Qwen2-VL-2B and Qwen2-VL-7B share architecture, connector, tokenizer, and training
recipe and differ only in scale, yet sit on opposite sides of the collapse on classification
(Table~\ref{tab:apppanel}). Re-running exactly this pair on recognized open-ended VQA benchmarks,
repurposed as many-shot ICL tasks, asks whether the prone/robust split survives the switch to an
open-ended format \emph{while architecture is held fixed}, so any surviving difference cannot be charged
to a connector or backbone confound. It survives (Table~\ref{tab:vqa}, Fig.~\ref{fig:vqa_curves}): the
\emph{same model ordering} carries over. The collapse-prone Qwen2-VL-2B degrades on VQAv2
($0.70\!\rightarrow\!0.62$, $\pm0.02$ over $3$ seeds) and TextVQA ($0.15\!\rightarrow\!0.11$), whereas the
robust Qwen2-VL-7B holds on both ($0.72\!\rightarrow\!0.74$; $0.20\!\rightarrow\!0.18$). The effect is
smaller in absolute terms here than on classification, for two reasons that are worth naming: VQA
soft-accuracy is bounded well below $1$ even zero-shot (there is less height to fall from), and open-ended
answers give the recency-copy heuristic less purchase than a closed $4$-way label set. \emph{Takeaway:
the collapse is a property of many-shot multimodal ICL in general, not of our synthetic protocol, and
which side of it a model falls on is the same on VQAv2 as on CIFAR, so the regime is a stable property of
the model rather than of the task.} The one apparent exception is
instructive: on ScienceQA both Qwen models \emph{improve} with shots ($0.18\!\rightarrow\!0.29$;
$0.21\!\rightarrow\!0.38$), because ScienceQA is a multiple-choice format whose demonstrations teach the
answer \emph{format} rather than fighting a perceptual prior, exactly the regime in which demonstrations
help. The collapse appears when demonstrations must override a strong pretraining prior, and recedes when
they instead supply a missing output convention.

\begin{table}[!ht]
\centering
\caption{The collapse on standard VQA benchmarks (VQA soft accuracy), $K{=}0\!\rightarrow\!K{=}16$
(mean over $3$ seeds; $95\%$ CI half-width $\le0.04$ on every cell). A within-family controlled pair
(same architecture, differing only in scale): the collapse-prone $2$B model degrades on the perceptual
benchmarks (VQAv2, TextVQA) while the robust $7$B model holds, reproducing their classification ordering
with architecture held fixed. Both \emph{rise} on ScienceQA, a multiple-choice task where demonstrations
supply an answer format rather than override a prior. All rows are means over $3$ seeds.}
\label{tab:vqa}
\begin{tabular}{lccc}
\toprule
Model & VQAv2 & TextVQA & ScienceQA \\
\midrule
Qwen2-VL-2B & $0.70\!\rightarrow\!0.62$ & $0.15\!\rightarrow\!0.11$ & $0.18\!\rightarrow\!0.29$ \\
Qwen2-VL-7B & $0.72\!\rightarrow\!0.74$ & $0.20\!\rightarrow\!0.18$ & $0.21\!\rightarrow\!0.39$ \\
\bottomrule
\end{tabular}
\end{table}

\begin{figure}[!ht]
\centering
\includegraphics[width=\textwidth]{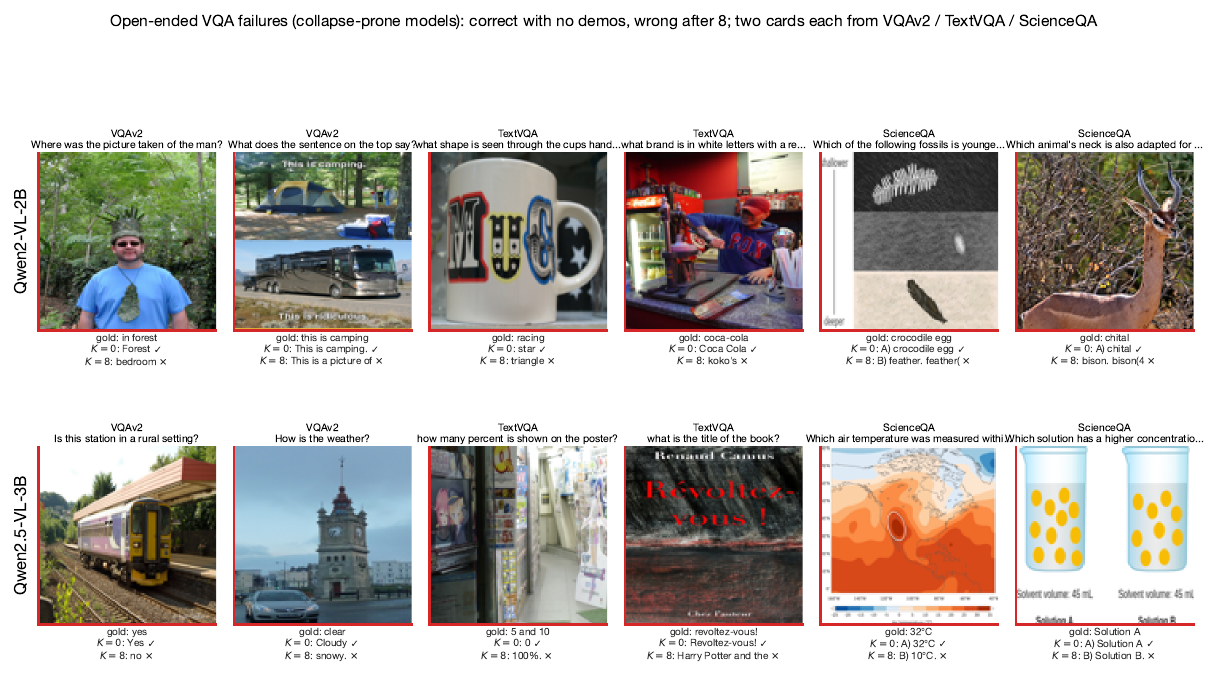}
\caption{\textbf{Open-ended VQA failures, query by query, for both collapse-prone models.} One row per
model (Qwen2-VL-2B, Qwen2.5-VL-3B); each row shows two real (image, question) pairs from each of VQAv2,
TextVQA, and ScienceQA, with the model's answer at $K{=}0$ (correct, $\checkmark$) and after eight
demonstrations ($K{=}8$, wrong, $\times$); gold answer shown. On both models and across all three
benchmarks the collapse turns correct, fluent answers into wrong ones (a correct fact replaced by a wrong
fact, or coherent text into a garbled phrase), mirroring the classification gallery
(Fig.~\ref{fig:gallery}) on open-ended tasks.}
\label{fig:vqa_failcase}
\end{figure}

The failures are interpretable at the level of individual questions, and the same pattern holds for
\emph{both} collapse-prone models on all three benchmarks (Fig.~\ref{fig:vqa_failcase}): a model that
answers an open-ended question correctly with no demonstrations gives a wrong, and sometimes degenerate,
answer once eight are prepended. On Qwen2-VL-2B, reading a brand correctly as ``Coca Cola'' at $K{=}0$
becomes ``koko's'' at $K{=}8$, and a fossil-age question answered ``crocodile egg'' becomes ``feather'';
on Qwen2.5-VL-3B, a book title read correctly as ``Revoltez-vous!'' becomes ``Harry Potter'', and a
temperature read ``$32^{\circ}$C'' becomes ``$10^{\circ}$C''. As in the synthetic case, the outputs stay
fluent; the failure is of judgement, not formatting.

\paragraph{What the collapse-prone models share.}
Taken together, the classification panel (Table~\ref{tab:apppanel}), the VQA benchmarks
(Table~\ref{tab:vqa}), and the frontier probe (Fig.~\ref{fig:frontier}) let us ask what distinguishes
the models that collapse from those that do not, across a deliberately heterogeneous set of
architectures. Three commonalities emerge. \emph{(i) It is the alignment recipe, not the connector
design or scale.} Collapse-prone models span MLP-projector (Qwen2-VL-2B, LLaVA-1.5/1.6), cross-attention
(Llama-3.2-11B), and high-token (LLaVA-OV-0.5B) designs, and range from $0.5$B to $11$B; collapse-resistant
models likewise span families and scales. Within a single family the divide is sharp and consistent (the
$2$B Qwen collapses, the $7$B does not), which points at the post-training that teaches a model how to
treat in-context demonstrations rather than at any structural property. \emph{(ii) The prone models share
a recency-copy signature.} Wherever the collapse appears, it is accompanied by the same behavioral
mechanism, a rising copy rate and \textsc{shuffled}$\approx$\textsc{standard} insensitivity to
demonstration content (Sec.~\ref{sec:dissoc}), so the prone models fail in the \emph{same way}: they
substitute ``answer like the most recent example'' for ``read the query.'' \emph{(iii) The collapse is
prior-interference, not task difficulty.} It is strongest exactly where demonstrations must override a
confident pretraining prior (natural-image classification, perceptual VQA) and recedes or reverses where
they instead supply a missing output convention (ScienceQA's multiple-choice format, on which even prone
models improve), and a frontier model that collapses on CIFAR is simultaneously perfect on the
prior-free shapes task. \emph{Takeaway: the collapse-prone models are not united by what they are built
from or how big they are, but by how they were aligned, and they all fail through the same recency-copy
shortcut triggered by prior-conflicting demonstrations, which is precisely the behavior the causal
analysis of Sec.~\ref{sec:circuit} localizes and the repair of Sec.~\ref{sec:resultsB} removes.}

\subsubsection{The collapse is a causally localized integration failure}\label{sec:circuit}
\begin{figure}[!ht]
\centering
\includegraphics[width=\textwidth]{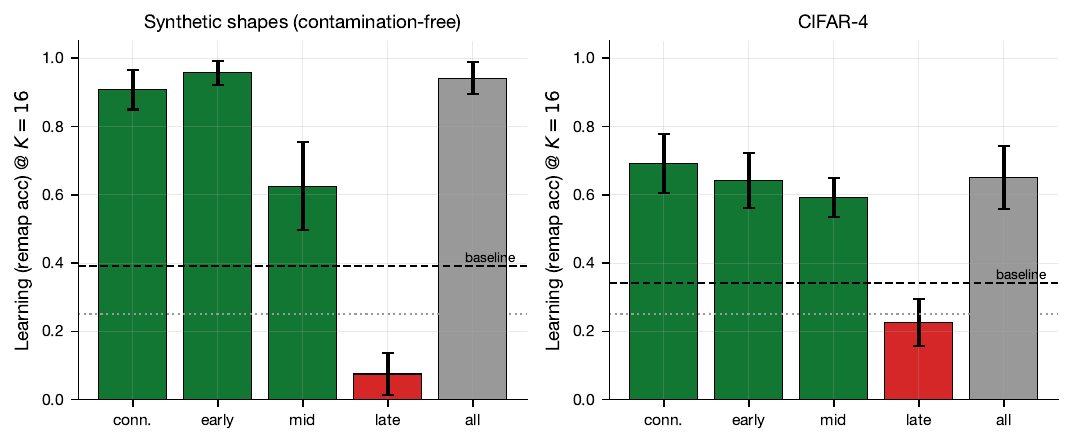}
\caption{\textbf{The collapse is causally removable at the integration interface, not the readout.}
In-context learning (\textsc{remap} accuracy at $K{=}16$; mean $\pm 95\%$ CI over $3$ seeds; chance
$=0.25$ dotted, baseline dashed) for Qwen2-VL-2B after a rank-$8$ adapter on each region.
Connector/early/mid (green) and the all-layer adapter restore genuine learning; the
\emph{equal-capacity} late readout adapter (red) does not, and on the contamination-free shapes task
drives learning below baseline. The connector, the smallest target, is the most consistent rescue
site.}
\label{fig:circuit}
\end{figure}

We now show the collapse is causally localized to the vision--language \emph{integration} pathway and is
separable from the \emph{readout}. We take a collapse-prone model (Qwen2-VL-2B), train a rank-$8$ adapter
restricted to a single region, and re-measure the \textsc{remap} K-curve: success means a model that
\emph{could not} learn the arbitrary mapping now can (Fig.~\ref{fig:circuit}, Table~\ref{tab:remove}).

\begin{table}[!ht]
\centering
\caption{The collapse is causally removable at the integration pathway but not the readout.
\textsc{remap} accuracy at $K{=}16$ (mean $\pm 95\%$ CI over $3$ seeds) for Qwen2-VL-2B after a
rank-$8$ adapter on each region; chance $=0.25$.}
\label{tab:remove}
\begin{tabular}{lcc}
\toprule
Adapter region & Shapes (remap@16) & CIFAR-4 (remap@16) \\
\midrule
none (baseline) & $0.39 \pm 0.05$ & $0.34 \pm 0.06$ \\
\textbf{connector} & $\mathbf{0.91 \pm 0.06}$ & $\mathbf{0.69 \pm 0.09}$ \\
early & $0.96 \pm 0.04$ & $0.64 \pm 0.08$ \\
mid & $0.63 \pm 0.13$ & $0.59 \pm 0.06$ \\
\textbf{late} & $\mathbf{0.08 \pm 0.06}$ & $\mathbf{0.23 \pm 0.07}$ \\
all & $0.94 \pm 0.05$ & $0.65 \pm 0.09$ \\
\bottomrule
\end{tabular}
\end{table}

\paragraph{Removable at the interface, not the readout.}
On the contamination-free shapes task, an adapter on the \emph{connector} lifts learning from $0.39$ to
$0.91$ and on \emph{early} layers to $0.96$, near ceiling, while restoring \textsc{standard} accuracy
to $\approx\!0.99$ (the collapse is removed). An \emph{equal-capacity} adapter on the \emph{late} layers
does the opposite, driving \textsc{remap} to $0.08$, \emph{below} the unintervened baseline. CIFAR-4
shows the same ordering (Table~\ref{tab:remove}). Because \emph{late} targets a parameter count
comparable to \emph{early}/\emph{mid} (the equal-capacity control of Sec.~\ref{sec:setup}: $70$ vs.\ $63$
adapter modules), the failure of the late intervention is not a capacity artifact; it is a statement
about \emph{where} the collapse lives. Three features of Table~\ref{tab:remove} deserve emphasis.

\emph{(i) A monotone front-to-back gradient.} Rescue efficacy falls steadily as the adapter moves deeper:
connector and early layers nearly saturate the task, mid is intermediate ($0.63$), and late not only
fails but actively suppresses learning. The collapse is thus not localized to a single module but to a
\emph{region}, the vision--language interface and the early language layers that first integrate the
demonstrations, with the readout playing no constructive role.

\emph{(ii) The smallest intervention is the most effective.} The connector is by far the lowest-parameter
target (a single projection block), yet it delivers the most consistent rescue across both tasks
(shapes $0.91$, CIFAR $0.69$), edging out the much larger all-layer adapter. This is strong evidence that
the collapse has a \emph{specific, compact} locus: adding capacity exactly at the modality bottleneck
buys more than adding capacity everywhere. \emph{Takeaway: you do not need to retrain the network to
remove the collapse, you need to touch the right $\sim\!1\%$ of it.}

\emph{(iii) The late adapter makes learning worse than doing nothing.} Driving \textsc{remap} from a
$0.39$ baseline down to $0.08$, well below chance, is not a null result; it means that spending capacity
on the readout while demonstrations are present actively reinforces the recency-copy heuristic rather
than the in-prompt rule. \emph{Takeaway: the readout is not merely the wrong place to help, it is a place
where ``adaptation'' actively entrenches the failure, which is why the locus of the fix is a substantive
claim and not a tuning detail.} Together these establish that the collapse is a deficit of the
\emph{integration} computation, surgically correctable at the interface, and it is this locus that
Sec.~\ref{sec:CircA} turns into a reusable repair.

\subsection{Resolving the collapse with \textsc{CircA}}\label{sec:resultsB}

\subsubsection{From mechanism to method: a transferable integration vaccine}\label{sec:CircA}

Sec.~\ref{sec:proposed} introduced \textsc{CircA} and its three components. We evaluate them across three
models, establishing first that the collapse is localized to the same integration pathway on every
architecture, then that a one-time \emph{vaccine} at that pathway repairs it on every architecture. The
\emph{vaccine} carries the load; the \emph{gate} and \emph{inject} components, examined afterward on the
original model, delimit where the repair must, and must not, live.
We select the three models on which the full causal pipeline (lesion-and-rescue plus vaccine) is run before
seeing any repair result, so that a positive finding cannot be the product of model search. They are
chosen to vary the two factors that could otherwise confound a structural claim: the \emph{connector
design} that bridges vision to language, and the \emph{language-model backbone}. Qwen2-VL-2B and
Qwen2.5-VL-3B are MLP-projector models but from different model generations; InternVL3-2B uses a
\emph{pixel-shuffle} connector and a different backbone entirely. Together they span the two dominant
connector families and two backbone lineages at a fixed small scale, so a shared integration-locus
mechanism cannot be attributed to a single projector type, tokenizer, or alignment recipe. All three are
collapse-prone yet retain enough competence for a rescue to be meaningful (Table~\ref{tab:apppanel}),
which is the regime in which the question is well-posed; deeply degenerate models, already at chance
before intervention, leave nothing for a small adapter to restore and are uninformative here. We report
all three together throughout this section rather than promoting one and replicating later.

\begin{figure}[!ht]
\centering
\includegraphics[width=\textwidth]{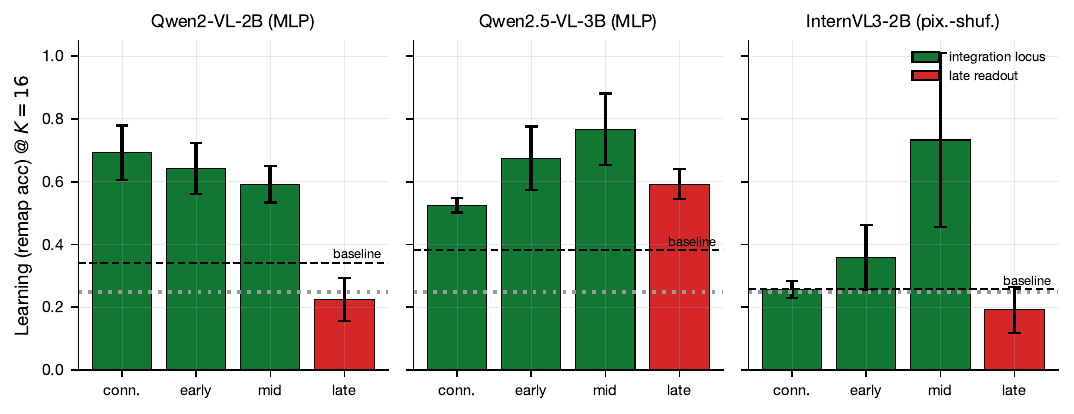}
\caption{\textbf{The lesion-and-rescue localizes the collapse across all three architectures.}
CIFAR-4 \textsc{remap} accuracy at $K{=}16$ (mean $\pm 95\%$ CI over $3$ seeds) after a rank-restricted
adapter on each region, for the two MLP-projector models (Qwen2-VL-2B, Qwen2.5-VL-3B) and the
pixel-shuffle InternVL3-2B. On every model the integration loci (connector / early / mid, teal) clear the
chance-level baseline (dashed) while the equal-capacity late readout (red) is the weakest region, at or
below baseline on Qwen2-VL-2B and InternVL3-2B and below the integration peak on Qwen2.5-VL-3B. The
integration-not-readout ordering is thus a property of collapse-prone VLMs rather than of one
architecture; InternVL3-2B is noisier across seeds.}
\label{fig:crossarch}
\end{figure}

\paragraph{The lesion-and-rescue localizes the collapse on every architecture.}
We run the same lesion-and-rescue on all three models on a common task (CIFAR-4 \textsc{remap}@$K{=}16$,
Fig.~\ref{fig:crossarch}): on each, an adapter at the \emph{integration} locus (connector / early / mid)
restores in-context learning while the equal-capacity \emph{late} readout adapter is the weakest region,
the same asymmetry established in Sec.~\ref{sec:circuit}. The ordering is sharp on the two MLP-projector
models, where the best integration locus roughly doubles the chance-level baseline (Qwen2-VL-2B: connector
$0.69$ vs.\ late $0.23$, at baseline; Qwen2.5-VL-3B: mid $0.77$ vs.\ late $0.59$), and reproduces on the
pixel-shuffle InternVL3-2B, where the late intervention sits at baseline (remap $0.19$) while a mid-layer
adapter lifts it to $0.73$. We note two honest caveats: on Qwen2.5-VL-3B the late adapter, though still the
weakest locus and well below the integration peak, does not collapse to baseline as it does on the other
two; and InternVL3-2B is noisier across seeds, so we read its positive side as a directional rather than a
clean point estimate. The integration-over-readout ordering nonetheless holds on all three, across two
connector families and two backbone lineages. This converges with
correlational interpretability: Kaduri et al.\ \cite{kaduri2024whatsimage} find cross-modal information
flow is dominated by the \emph{middle} layers, with early and late layers contributing only marginally,
the very layers our intervention identifies as the editable integration site; what that work locates by
observation, the lesion-and-rescue confirms by manipulation. It also refines the finding that visual
information becomes linearly \emph{readable} only in late layers
\cite{neo2024interpvisual,basu2024entitymodal}: the late layers \emph{read out} the decision but do not
\emph{integrate} the demonstrations, so readability and manipulability dissociate.

\begin{figure}[!ht]
\centering
\includegraphics[width=\textwidth]{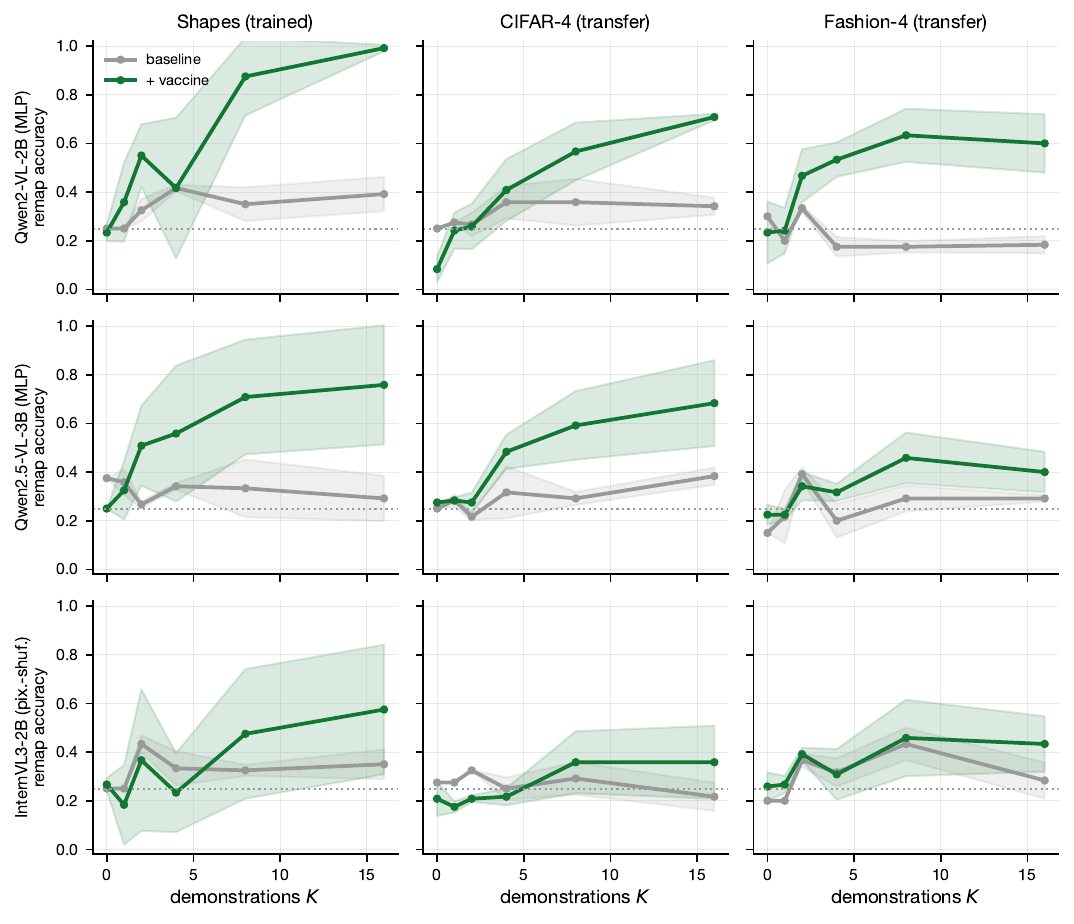}
\caption{\textbf{A one-time integration vaccine confers transferable collapse-resistance across
architectures.} \textsc{remap} accuracy versus $K$ for each model (rows) before (grey) and after (teal) a
single integration-locus adapter trained \emph{only} on shapes remap (mean over $3$ seeds, $95\%$ CI
bands; chance $=0.25$, dotted). On every architecture the vaccine climbs with $K$ on the trained task
(left) and lifts two \emph{unseen} task families (middle, right) from chance, while the baseline stays
pinned at chance throughout. The effect is largest on the MLP-projector models and more modest, though
still positive on both held-out tasks, on the pixel-shuffle InternVL3-2B.}
\label{fig:vaccine}
\end{figure}

\paragraph{A one-time vaccine transfers to unseen tasks.}
We then train a single rank-restricted adapter on the connector and early--mid layers using \emph{only}
the shapes remap task, freeze it, and evaluate \textsc{remap} accuracy with no further training on shapes
and on two held-out task families (Figure~\ref{fig:vaccine}, Table~\ref{tab:vaccine}). The pattern is the
same on all three models: the un-vaccinated model sits at chance across all $K$ on every task (the
collapse), while the vaccinated model climbs with $K$ on the trained task and, crucially, on two families
the adapter never saw. At $K{=}16$ the vaccine lifts the trained shapes task from chance to
$0.99/0.76/0.58$ (Qwen2-VL-2B / Qwen2.5-VL-3B / InternVL3-2B) and the held-out families from chance to
$0.71/0.68/0.36$ on CIFAR-4 and $0.60/0.40/0.43$ on Fashion-4. Because resistance generalizes to tasks
outside the training distribution, the vaccine installs a general ``use the in-context mapping''
capability at the integration pathway rather than a task-specific lookup.

\begin{table}[!ht]
\centering
\caption{\textbf{The integration vaccine, trained once on shapes remap, transfers to unseen task families
across architectures.} \textsc{remap} accuracy at $K{=}16$ (mean over $3$ seeds, $n{=}40$ held-out queries
per seed; chance $=0.25$); each cell is baseline\,$\rightarrow$\,+vaccine. The localization and the
transferable repair hold for an MLP-projector model, a later-generation MLP-projector model, and a
pixel-shuffle model.}
\label{tab:vaccine}
\setlength{\tabcolsep}{4pt}
\begin{tabular}{lccc}
\toprule
Model & Shapes$^\dagger$ & CIFAR-4 & Fashion-4 \\
 & (trained) & (transfer) & (transfer) \\
\midrule
Qwen2-VL-2B (MLP) & $0.39\!\rightarrow\!\mathbf{0.99}$ & $0.34\!\rightarrow\!\mathbf{0.71}$ & $0.18\!\rightarrow\!\mathbf{0.60}$ \\
Qwen2.5-VL-3B (MLP) & $0.29\!\rightarrow\!\mathbf{0.76}$ & $0.38\!\rightarrow\!\mathbf{0.68}$ & $0.29\!\rightarrow\!\mathbf{0.40}$ \\
InternVL3-2B (pix.-shuf.) & $0.35\!\rightarrow\!\mathbf{0.58}$ & $0.22\!\rightarrow\!\mathbf{0.36}$ & $0.28\!\rightarrow\!\mathbf{0.43}$ \\
\bottomrule
\end{tabular}

\vspace{2pt}
{\footnotesize $^\dagger$ each cell: baseline\,$\rightarrow$\,+vaccine \textsc{remap} accuracy at $K{=}16$.}
\end{table}

\paragraph{Reading the cross-model pattern.}
Three regularities emerge from comparing the models. First, the \emph{direction} is invariant: on every
architecture the vaccine lifts both the trained task and the two unseen families above chance, and the
baseline never does, so the repair is a property of collapse-prone VLMs rather than of one model.
Second, the \emph{magnitude} tracks the connector family: the effect is largest on the two MLP-projector
models (held-out gains of $+0.37/+0.42$ and $+0.30/+0.11$ on CIFAR/Fashion) and more modest on the
pixel-shuffle InternVL3-2B ($+0.14/+0.15$), mirroring the noisier lesion signal there and suggesting that
a simple linear projector exposes a cleaner integration locus for a low-rank edit than a spatially
reshuffled one. Third, the gain is monotone in $K$ on all three models, concentrated at the larger
demonstration counts where the collapse is deepest (Fig.~\ref{fig:vaccine}), so the vaccine repairs
exactly the many-shot regime it targets rather than shifting the whole curve uniformly. Together these
say the integration locus is the same editable site on connectors and backbones that otherwise share no
implementation, and that the headroom for a low-rank repair is set by how cleanly that site is exposed.

\paragraph{The fix must live at the integration locus.}
Ablating the vaccine confirms the integration pathway is the causal site (Table~\ref{tab:CircA}).
Re-targeting the same-capacity adapter to the \emph{late} layers, the consolidation site, not only
fails to transfer but collapses even the trained task to $0.00$, mirroring the late-adapter failure in
the diagnostic lesion (Table~\ref{tab:remove}). Within the integration locus the components are
partially redundant for the easier CIFAR transfer (every single-component drop still transfers, CIFAR
$\ge0.64$ at $K{=}16$) but more complementary for the harder Fashion transfer, which erodes as the early
or mid language layers are removed ($0.60\!\rightarrow\!0.52/0.53$); the connector is the least
load-bearing. The transferable competence is carried by the early--mid language layers, the same locus
the lesion study identified, not by the connector alone and not by the readout.

\begin{table}[!ht]
\centering
\caption{\textbf{Locus ablation of the vaccine.} \textsc{remap} accuracy at $K{=}16$ (Qwen2-VL-2B,
mean $\pm 95\%$ CI over $3$ seeds, $n{=}40$ per seed; chance $=0.25$) for a single adapter trained on
shapes remap and evaluated on the trained task and two held-out families. The repair requires the
integration locus: the equal-capacity \emph{late} variant collapses, while integration-locus variants
transfer.}
\label{tab:CircA}
\begin{tabular}{lccc}
\toprule
Vaccine locus & Shapes (trained) & CIFAR-4 (transfer) & Fashion-4 (transfer) \\
\midrule
none (baseline) & $0.39 \pm 0.07$ & $0.34 \pm 0.04$ & $0.18 \pm 0.04$ \\
\textbf{full} (conn+early+mid) & $\mathbf{0.99 \pm 0.01}$ & $\mathbf{0.71 \pm 0.01}$ & $\mathbf{0.60 \pm 0.12}$ \\
\quad$-$connector & $1.00 \pm 0.00$ & $0.72 \pm 0.07$ & $0.58 \pm 0.09$ \\
\quad$-$early & $0.83 \pm 0.12$ & $0.64 \pm 0.04$ & $0.52 \pm 0.11$ \\
\quad$-$mid & $1.00 \pm 0.00$ & $0.70 \pm 0.14$ & $0.53 \pm 0.07$ \\
\textbf{late only} & $\mathbf{0.00 \pm 0.00}$ & $\mathbf{0.31 \pm 0.05}$ & $\mathbf{0.37 \pm 0.09}$ \\
\bottomrule
\end{tabular}
\end{table}

\paragraph{Training-free injection does not fix the collapse, and that is itself informative.}
Task and function vectors transplant a computation a model \emph{already performs} into a cheaper form
\cite{hendel2023taskvectors,todd2024functionvectors,huang2024mmtaskvectors}; our inject path is their
multimodal analogue, extracting the task vector from the model's \emph{own} forward pass and adding it
at the integration locus. On the collapse-prone model it leaves \textsc{remap} accuracy at chance at
both the early and mid loci, identical to the zero-shot baseline. Far from a null result, this pins
down the nature of the collapse: a training-free patch can only relocate a computation that exists, so
its failure shows the collapse is the \emph{absence} of the integration computation rather than its
misplacement. The fix must therefore \emph{add} capacity at the integration locus (the vaccine),
consistent with the equal-capacity late adapter failing to rescue the collapse (Table~\ref{tab:remove}).
Injection remains useful for context amortization once a model can integrate (post-vaccine or
collapse-resistant), which the gate exploits at serving time.

\paragraph{The repair strategies, head to head.}
Table~\ref{tab:baseline} places the three \textsc{CircA} components against the two naive baselines in a
single transfer setting (train on shapes remap, evaluate \textsc{remap} on the unseen CIFAR-4 at
$K{=}16$). The ranking is unambiguous: only the vaccine clears the collapse ($0.78\pm0.09$), more than
doubling the next-best strategy. Simply adding more demonstrations (\emph{many-shot}, $0.38$) is the
collapse itself; training-free \emph{injection} ($0.29$) and the runtime \emph{gate} ($0.21$) sit at or
below the zero-shot prior ($0.25$), confirming that neither relocating an absent computation nor gating
the context substitutes for adding integration capacity. The comparison is between repair \emph{strategies}
and naive controls on the same model, not against an external method; we are not aware of a published
multimodal many-shot repair baseline to compare against, which is itself indicative of the gap this work
addresses.

\begin{table}[!ht]
\centering
\caption{\textbf{Repair strategies versus naive baselines}, all on the collapse-prone Qwen2-VL-2B in one
transfer setting: trained on shapes remap, evaluated on held-out CIFAR-4 \textsc{remap} at $K{=}16$
(mean $\pm 95\%$ CI over $3$ seeds; chance $=0.25$). Only the integration vaccine clears the collapse.}
\label{tab:baseline}
\begin{tabular}{lc}
\toprule
Strategy & CIFAR-4 \textsc{remap} @$K{=}16$ \\
\midrule
zero-shot (prior) & $0.25 \pm 0.00$ \\
many-shot (the collapse) & $0.38 \pm 0.07$ \\
inject (training-free task vector) & $0.29 \pm 0.01$ \\
gate (runtime context gating) & $0.21 \pm 0.04$ \\
\textbf{vaccine (integration adapter)} & $\mathbf{0.78 \pm 0.09}$ \\
\bottomrule
\end{tabular}
\end{table}

\subsection{Analytic and Ablation Experiments}\label{sec:resultsC}

Having established the collapse, its locus, and its repair across the three architectures, we now report
finer-grained analyses on the original model that probe the mechanism and its boundaries: how the rescue
depends on the demonstration count, how it splits across attention and MLP, and how the integration locus
relates to where durable memories are best written.

\subsubsection{K-specificity of the rescue}
The rescue is concentrated where the collapse is. At $K{\le}2$ the integration and late interventions are
indistinguishable from baseline; the separation emerges at $K{=}4$ and is maximal at $K{=}8,16$. On
shapes, connector \textsc{remap} rises
$0.38\!\rightarrow\!0.73\!\rightarrow\!0.90\!\rightarrow\!0.91$ across $K{=}2,4,8,16$, while the late
adapter stays at the floor ($0.31\!\rightarrow\!0.05\!\rightarrow\!0.07\!\rightarrow\!0.08$); the full
curves are in Table~\ref{tab:appcurves}. The effect tracks the demonstration-overload regime rather
than a uniform change in competence.

\begin{table}[!ht]
\centering
\caption{\textbf{Full lesion-and-rescue K-curves} (\textsc{remap} accuracy, Qwen2-VL-2B, mean over $3$
seeds) after a rank-$8$ adapter on each region; chance $=0.25$. Connector/early/mid/all rise with $K$;
the equal-capacity \emph{late} adapter stays at or below chance, and the integration--late gap opens
only once demonstrations accumulate ($K\ge4$).}
\label{tab:appcurves}
\begin{tabular}{llcccccc}
\toprule
Task & Region & $K{=}0$ & $K{=}1$ & $K{=}2$ & $K{=}4$ & $K{=}8$ & $K{=}16$ \\
\midrule
\multirow{6}{*}{Shapes}
 & baseline & 0.25 & 0.25 & 0.32 & 0.42 & 0.37 & 0.39 \\
 & connector & 0.18 & 0.28 & 0.38 & 0.73 & 0.90 & 0.91 \\
 & early & 0.12 & 0.31 & 0.46 & 0.78 & 0.91 & 0.96 \\
 & mid & 0.19 & 0.40 & 0.38 & 0.18 & 0.35 & 0.63 \\
 & late & 0.00 & 0.25 & 0.31 & 0.05 & 0.07 & 0.08 \\
 & all & 0.14 & 0.29 & 0.46 & 0.72 & 0.90 & 0.94 \\
\midrule
\multirow{6}{*}{CIFAR-4}
 & baseline & 0.25 & 0.28 & 0.28 & 0.38 & 0.35 & 0.34 \\
 & connector & 0.25 & 0.23 & 0.22 & 0.52 & 0.66 & 0.69 \\
 & early & 0.23 & 0.38 & 0.25 & 0.49 & 0.53 & 0.64 \\
 & mid & 0.28 & 0.32 & 0.23 & 0.46 & 0.50 & 0.59 \\
 & late & 0.18 & 0.28 & 0.30 & 0.39 & 0.29 & 0.23 \\
 & all & 0.25 & 0.40 & 0.24 & 0.49 & 0.54 & 0.65 \\
\bottomrule
\end{tabular}
\end{table}

\subsubsection{Attention and MLP both carry the rescue}
\begin{figure}[!ht]
\centering
\includegraphics[width=0.6\textwidth]{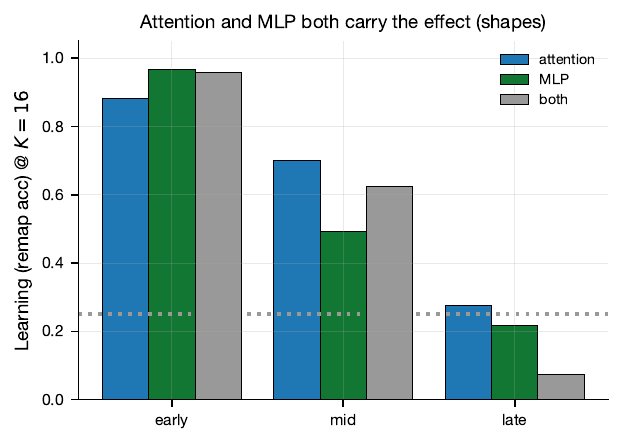}
\caption{\textbf{Attention and MLP both carry the rescue.} \textsc{remap} accuracy at $K{=}16$
(Qwen2-VL-2B, shapes) for adapters restricted to attention-only, MLP-only, or both, within the early,
mid and late regions. The integration effect is distributed across attention and MLP at the early/mid
interface; neither sub-module rescues learning in the late readout (chance $=0.25$, dotted).}
\label{fig:component}
\end{figure}

Splitting each region into attention-only and MLP-only adapters shows the effect is distributed, not
isolated to a single sub-module. On shapes (remap@$16$), an early-layer MLP-only adapter reaches $0.97$
and attention-only $0.88$; a mid-layer adapter favors attention ($0.70$ vs.\ $0.49$); in the late
layers neither rescues learning (attention $0.28$, MLP $0.22$; Fig.~\ref{fig:component}). CIFAR-4 shows the same ordering.

\subsubsection{Integration is not consolidation}\label{sec:consol}
Because in-context adaptation is weight-free, the collapse is a capacity bound on the fast path: past it,
durable adaptation must be written into weights. Is the same integration circuit that fixes the collapse
also the right place to consolidate? We test this on a $5$-task class-incremental stream (CIFAR-100
splits) by restricting a sequential LoRA to three loci, the \emph{circuit} (connector\,+\,early/mid),
\emph{all} layers, and \emph{late} layers, and measuring average accuracy (AA) and backward transfer
(BWT).

\begin{figure}[!ht]
\centering
\includegraphics[width=0.85\textwidth]{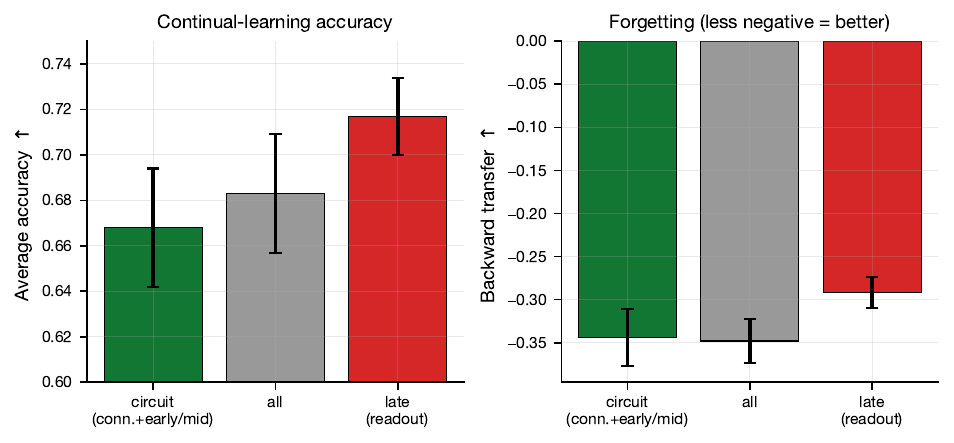}
\caption{\textbf{Integration $\neq$ consolidation.} Continual learning over a $5$-task
class-incremental stream (mean $\pm$ s.d.\ over $3$ seeds), consolidating with a LoRA restricted to each
locus. The \emph{late} readout (red) consolidates best, highest average accuracy (left) and least
forgetting (right), at the fewest parameters, \emph{not} the early/mid integration circuit (green) that
removes the collapse in Fig.~\ref{fig:circuit}.}
\label{fig:consol}
\end{figure}

The answer is no, and informatively so (Fig.~\ref{fig:consol}). Consolidation is most durable at the
\emph{late} layers: they reach the highest average accuracy ($0.717$ vs.\ $0.668$ for the circuit), the
least forgetting ($-0.292$ vs.\ $-0.344$ BWT), and do so with the \emph{fewest} trainable parameters
($3.0$M vs.\ $6.4$M). The locus of in-context integration is thus \emph{distinct} from the locus of
effective weight-based consolidation: the early/mid interface fixes the fast path, while the late
readout is the efficient home for the slow path. This is a second dissociation, and a concrete design
rule: ride ICL until the collapse, then consolidate at the readout, not at the integration circuit.

\section{Conclusion}\label{sec:conclusion}
Many-shot multimodal in-context learning is not monotonically helpful: it exhibits a graded
\emph{collapse} that, surprisingly, reaches frontier scale. The collapse exposes a dissociation between
robustness and genuine learning, and it is causally localized to the vision--language integration
pathway, removable at the connector and early/mid layers but not at the late readout. That same
readout, not the integration circuit, is where weight-based consolidation is most durable. The
in-context collapse is therefore best understood as an integration failure at the modality boundary,
mechanistically distinct from both the readout and the consolidation locus, and it provides a concrete
diagnostic for when a vision-language model must stop prompting and start consolidating.

The collapse is an in-context \emph{integration} failure: demonstrations overload the vision--language
interface and displace, rather than refine, the model's prior. It is distinct from catastrophic
forgetting, a weight-overwrite failure; the bridge between the two is architectural rather than an
equivalence, in that the collapse bounds the weight-free path and thereby motivates the weight-based one.
Practically, because the collapse is localized and the fix transfers, immunity can be installed offline: a
single small adapter on the integration locus, an integration \emph{vaccine}, confers collapse-resistance
that generalizes to unseen task families without per-task training (\S\ref{sec:CircA}), turning the
diagnosis into a cheap, deployable intervention.

Two boundaries of the present account deserve emphasis. First, the intervention is \emph{asymmetric}:
removing the collapse in a collapse-prone model is clean, but the reverse, \emph{inducing} a collapse in a
robust learner by lesioning the same circuit, is weak and diffuse in our experiments. This asymmetry
suggests collapse-\emph{immunity} is a distributed property not reducible to a single circuit, whereas
collapse-\emph{proneness} is a localized integration deficit that a small adapter can correct, and it
means our causal claim is best supported in the prone-to-immune direction. Second, the repair was
validated on three small open VLMs spanning two connector families and two backbone lineages; the frontier
evidence (\S\ref{sec:dissoc}) establishes that the \emph{phenomenon} scales, but whether the same
low-rank integration edit repairs proprietary frontier models, which we can probe only through inference
APIs (\S\ref{sec:dissoc}), remains open.

Three directions strike us as both unexplored and high-leverage. (i)~\emph{A collapse predictor from
pretraining signatures.} The integration locus is identifiable by intervention; if the same locus leaves a
\emph{measurable} fingerprint, for instance the rank or conditioning of the connector-to-early-layer
Jacobian, one could predict a model's collapse-proneness directly from its weights and pre-vaccinate at
release time, without ever running the many-shot probe. (ii)~\emph{Closing the prompt--consolidate loop at
serving time.} We show \emph{when} a model should stop prompting and start consolidating, but not yet how
to do so online; an agent that monitors the copy-rate gate, triggers a lightweight integration update when
the gate saturates, and folds the just-seen demonstrations into the vaccine would turn the static
diagnostic into a continually adapting controller. (iii)~\emph{Native multimodal and interleaved
demonstrations.} Our probes place one image per demonstration in a single turn; extending the analysis to
interleaved image--text sequences, video frames, and audio would test whether the integration locus is a
property of the vision--language boundary specifically or a general signature of any modality fusion under
many-shot pressure, the more ambitious structural claim.

\section*{Data and code availability}
Panel configurations, the synthetic-concept generator, the intervention harness, and the result files
behind every figure and table are released with the paper to enable full reproduction: \url{https://github.com/rostami-m/incontext-collapse}.

\bibliography{references}

\end{document}